# A machine learning-based viscoelastic-viscoplastic model for epoxy nanocomposites with moisture content

Betim Bahtiri[a,∗], Behrouz Arash[b], Sven Scheffler[a], Maximilian Jux[c], Raimund Rolfes[a]

[a]*Institute of Structural Analysis, Leibniz Universität Hannover, Appelstraße 9A, 30167 Hannover, Germany*
[b]*Department of Mechanical, Electronic and Chemical Engineering, OsloMet, Pilestredet 35, Oslo 0166, Norway*
[c]*Institute of Lightweight Systems, Multifunctional Materials, DLR (German Aerospace Center), Lilienthalplatz 7, 38108 Brunswick, Germany*

**Abstract**

In this work, we propose a deep learning (DL)-based constitutive model for investigating the cyclic viscoelastic-viscoplastic-damage behavior of nanoparticle/epoxy nanocomposites with moisture content. For this, a long short-term memory network is trained using a combined framework of a sampling technique and a perturbation method. The training framework, along with the training data generated by an experimentally validated viscoelastic-viscoplastic model, enables the DL model to accurately capture the rate-dependent stress-strain relationship and consistent tangent moduli. In addition, the DL-based constitutive model is implemented into finite element analysis. Finite element simulations are performed to study the effect of load rate and moisture content on the force-displacement response of nanoparticle/epoxy samples. Numerical examples show that the computational efficiency of the DL model depends on the loading condition and is significantly higher than the conventional constitutive model. Furthermore, comparing numerical results and experimental data demonstrates good agreement with different nanoparticle and moisture contents.

## 1. Introduction

Innovative designs of polymer nanocomposites have led to the development of advanced materials with complex material response [1]. To accurately predict the history-dependent behavior of the materials under hygrothermal conditions, more complex constitutive models with additional parameters have been developed in the literature [2, 3]. Although the proposed constitutive models are able to describe the highly nonlinear behavior of the materials, a major challenge is to reduce the error between the numerical predictions and the experimental data under complex loading conditions. In addition, since the numerical modeling of the nonlinear material behavior involves time-consuming iterative solutions, the ability of deep learning (DL)-based constitutive models to improve the computational efficiency of the numerical methods has attracted attention [4–8].

---

∗Corresponding author
*Email address:* b.bahtiri@isd.uni-hannover.de (Betim Bahtiri)

One of the first works in this direction is done by Ghaboussi et al. [9] presenting a neural network approach to unify the mechanical behavior of plain concrete directly from experimental data. Since the material behavior is stated to be path-dependent, they trained the neural network to predict the strain increments given the current state of stress, strain and stress increment. Due to the requirement of a comprehensive set of experiments, the model was able to predict only biaxial and uniaxial loading. This work is extended to an auto-progressive training and therefore reduction of needed experimental data [10]. Stoffel et al. [11] utilized the nonlinear stress-strain behavior of an aluminium tube under pressure to train a neural network and substitute a viscoplastic model within a finite element (FE) framework. Since the neural network applied is not able to intrinsically learn the path-dependent behavior of the viscoplastic model they included the plastic strain tensor and backstress tensor as additional inputs to predict the path-dependency. They showed that the computational time using the neural network was up to 50% lower than in the case of the FE simulation. Similar efforts to replace the constitutive model were presented in [12], where the gated recurrent unit structure together with an attention mechanism is incorporated to learn the history-dependent elasto-plastic mechanical behavior of low-yield-point steels under cyclic loading conditions. They were able to depict the nonlinear behavior of the constitutive model and underlined the importance of the loss-function used in the learning process of neural networks. Recently, long short-term memory Networks (LSTM), which belong to the class of recurrent neural networks (RNNs), have been incorporated with a $FE^2$ [13] computational homogenization technique to develop an anisotropic plasticity model for heterogeneous material under arbitrary loading paths [14]. The authors map the strain tensor, the material parameters and an averaged strain to depict the stress tensor by creating 14,000 sets of data to train the LSTM. The proposed LSTM network is proving to be very effective in capturing the arbitrary loading paths for the two-dimensional case. A similar approach is provided by Ghavamian et. al [15] who proposed a strategy to collect stress-strain data from the micromechanical models of academic examples using a viscoplastic constitutive model and implement it into the FE$^2$ homogenization scheme. The authors utilize the automatic differentiation of RNNs to compute the consistent tangent tensor.

The above mentioned RNNs are developed by introducing recursions into the data flow and therefore learning from the previous time step [16, 17]. This allows the neural networks to learn long-term dependencies between timesteps of a sequence data and has been largely used in DL for classification or regression tasks [18, 19]. However, Hochreiter et al. [20] reported that RNNs suffer from fading memory and vanishing or exploding gradient, which motivated the development of LSTMs by introducing gating of the architecture [21]. Through the gating approach, the network is able to bridge minimal time legs by enforcing a constant error flow through a error carousel within the units of the LSTM. Therefore, LSTM networks control the data flow and address the fading memory and vanishing or exploding gradient problems associated to regular RNNs [22]. This makes the LSTM a powerful network to depict history-dependent and nonlinear mechanical behavior [23, 24].

Besides the increase in computational efficiency, DL models are also constructed to enhance computational mechanics in a different manner. Sadeghi and Lotfan [25] proposed an approach to implement neural networks for system identification and parameter estimation for a cantilever beam model in the presence of noise. Since a closed form for estimating the non-linear parameters is not available, the authors introduce a neural network approach. The results are validated using cross-validation, and the corresponding outputs confirm the predictability of the approach. Koeppe et al. [26] presented "meta elements," which maintain the main idea of FE but lead to a significant model order reduction. The model predicts field variables and forces using neural networks while reducing the number of degrees of freedom. A similar approach is followed by Tandale et al. [27], who incorporate LSTM networks within the finite element framework to predict the equivalent plastic strain of the elasto-viscoplastic constitutive model. The authors apply a self-learning approach to make the neural network adaptable to different materials by introducing a modified loss function. Other works extend the usage of the DL models within computational mechanics by implementing physics-informed neural networks, which learn the solution of the partial differential equations by utilizing a loss function compromised of the residual error of the linear equilibrium and its boundary conditions [28–31].

Although DL models have been developed for traditional materials, the development of material models for new high-performance materials, such as polymer nanocomposites, is of great significance. Recently, nanoparticle-reinforced thermosets have shown to be a promising composite material, where the low weight of epoxy resins is combined with the nanoparticles' features [1]. Among different particles, boehmite nanoparticles (BNPs) have been considered to improve the material properties of the composites, including shear strength, compressive strength, and fracture toughness [2, 32]. Furthermore, intensive research activities on constitutive models for composites have led to various models trying to reproduce the nonlinear rate-dependent mechanical behavior of the material [33–36].

Among others, Boyce et al. [37] developed a finite deformation-based model for thermoplastic polyurethanes exhibiting strong hysteresis and softening. The material is composed of hard and soft segments to take into account in the constitutive model derivation. Melro et al. [38] introduced a constitutive model for continuous-fiber reinforced composites with a fixed strain rate and therefore ignoring the viscoelasticity. The authors presented a model based on a paraboloidal yield criterion to separate yield strengths under tension or compression and pressure sensitivity. Poulain et al. [39] developed a finite deformation constitutive model for epoxy resins and implemented a modified argon model [40] to predict the viscoelastic behavior depending on temperature. They calibrated the model over a wide range of temperatures and strain rates for uniaxial loading conditions. Arash et al. [2] developed a physically informed model at finite deformations by utilizing molecular dynamics simulations [41, 42] to predict material parameters and validated the model against uniaxial loading conditions. The study shows that introducing atomistic simulations can reduce the number of experiments required to calibrate the material model. They also successfully validated the model for uniaxial loading. Recently, Arash et al. [3] developed a phase-field fracture model for polymer nanocomposites and incorporated the moisture dependency in the constitutive model. Their results show good agreement with experimental data at different nanoparticle contents. Regarding the modeling of cyclic behavior, Rocha et al. [43] recently developed a model based on small strains and included the moisture dependency for an epoxy system. They were able to predict several phenomena, such as nonlinear reloading branches. Still, they could not accurately estimate the amount of plastic strain. Silberstein et al. [36] presented a viscoelastic-viscoplastic model for electrolyte membrane Nafion at ambient conditions by introducing a back stress approach leading to accurate results for the mentioned material system.

Based on these investigations from the literature, some unaddressed issues still need to be addressed. Firstly, the models are usually calibrated against uniaxial loading conditions, which is insufficient to capture the viscoelastic behavior of polymer-based materials fully. Secondly, the constitutive models are mainly developed for neat polymeric material or composites at dry conditions by neglecting ambient conditions and the influence of moisture content in combination with particle contents. Thirdly, the models mostly fail to capture the time-dependent irreversible response fully, or the predictions must be more accurate.

To address the open questions, we propose several additions to improve the performance of the constitutive model for a nanoparticle reinforced and highly crosslinked thermosetting polymer with moisture content. First, to capture the highly nonlinear hysteresis behavior, we propose a nonlinear and physically motivated viscoelastic model, which is modified to capture the strong hysteresis of the nanocomposite. Secondly, a stress softening damage approach is implemented to the constitutive model to model the quasi-irreversible sliding of the molecular chains. Finally, an additional viscoplastic dashpot, including strain-rate dependency, is added to the constitutive model to depict the time-dependent irreversible viscoplastic response. The model is derived for investigating the effect of moisture content on the stress-strain behavior of BNP/epoxy nanocomposite at finite deformation. Due to the complexity of the constitutive model, a DL model is developed to increase the computational efficiency of the numerical simulations.

The present work is organized as follows. First, we present the nonlinear viscoelastic-viscoplastic damage model in Section 2, including moisture and nanoparticle volume fraction dependency at finite deformations. The integration and explanation of the multilayered LSTM network are then provided in Section 3. Here, insights are delivered into the training technique to predict the constitutive model and the tangent modulus tensor needed in the finite element framework. Next, a finite element formulation is presented in Section 4 to integrate our constitutive- and DL model. In Section 5, the proposed constitutive model is validated using numerical simulations, and the effect of moisture, BNPs volume fraction on the cyclic loading-unloading behavior of the nanocomposites is investigated. Beyond this, we underline the benefits of the DL model compared with the conventional constitutive model by highlighting the efficiency increase. The study may broadly impact the usage of DL models in computational mechanics in the sense that the proposed model can replace a complex and highly time-consuming time integration of the constitutive model yet keep the advanced predictability of the material model.

## 2. Constitutive model for nanoparticle/epoxy

In this section, a viscoelastic-viscoplastic damage model for BNP/epoxy nanocomposites is proposed. The stress response is decomposed into an equilibrium part and two viscous parts to capture the nonlinear rate-dependent behavior of the materials. The effect of BNPs and moisture on the stress-strain relationship is taken into account by defining an amplification factor as a function of the nanofiller and moisture contents. Here, we also take into account the material swelling through moisture. The proposed model is an extension of previous work by Bergstrom and Hilbert [44], aimed at predicting the experimentally observed response of BNP/epoxy nanocomposites under cyclic loading.

*2.1. Kinematics*

The total deformation gradient, containing the mechanical deformation, is multiplicatively split into a volumetric and deviatoric part as

$$\mathbf{F} = J^{1/3}\,\mathbf{F}_{iso}, \tag{1}$$

where $J = \det[\mathbf{F}]$ and $\mathbf{F}_{iso}$ are the volumetric deformation and the isochoric deformation gradient, respectively. The volume deformation is further decomposed into two terms: The mechanical compressibility $J_m$ and the moisture-induced swelling $J_w$, leading to an overall volumetric deformation as

$$J = J_m\,J_w, \tag{2}$$

where

$$J_w = 1 + \alpha_w\,w_w. \tag{3}$$

In the equation above, $\alpha_w$ is the moisture swelling coefficient and $w_w$ is the moisture content [3, 45]. Our model incorporates experimental characteristics by decomposing the material behavior into a viscoelastic and a viscoplastic part, corresponding to the time-dependent reversible and time-dependent irreversible response, respectively. We further decompose the viscoelastic stress response into a hyperelastic network and a viscous network. The hyperelastic spring, associated with the entropy change due to deformations, captures the equilibrium response, while the viscous network composed of an elastic spring and a viscoelastic dashpot describes the non-equilibrium behavior of the nanocomposites. Additionally, the quasi-irreversible sliding of the molecular chains, resulting in stress softening, also known as the Mullins effect [46, 47], is implemented within the constitutive model. A schematic structure of the model is presented in Fig. 1.

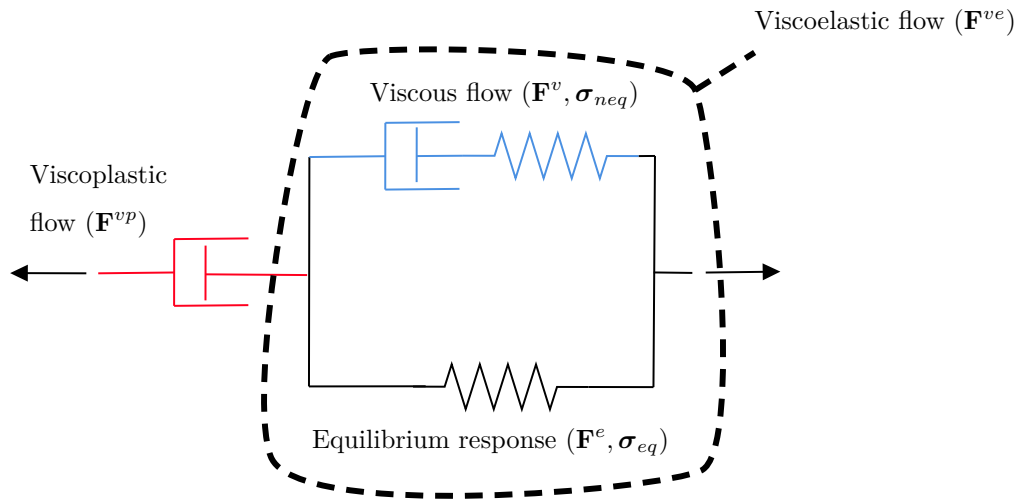

**Figure 1:** One-dimensional schematic of the viscoelastic-viscoplastic constitutive model.

The proposed constitutive model is able to capture the following main features of the material behavior:

1. Nonlinear elasticitiy at finite deformation;
2. nonlinear viscoelastic behavior;
3. viscoplastic flow because of stress driven chain sliding;
4. stress softening during deformation; and
5. the effect of moisture content on the stress-strain relationship.

The deviatoric part of the deformation gradient is decomposed into a viscoplastic and a viscoelastic component [48]:

$$\mathbf{F}_{iso} = \mathbf{F}^{ve}_{iso}\mathbf{F}^{vp}_{iso}. \qquad (4)$$

Also, the viscoelastic deformation gradient is split into an elastic and an inelastic part as

$$\mathbf{F}^{ve}_{iso} = \mathbf{F}^{e}_{iso}\mathbf{F}^{v}_{iso}. \qquad (5)$$

Accordingly, similar decompositions are obtained for the left Cauchy-Green deformation ten-

sors:

$$\mathbf{B}_{iso} = \mathbf{F}_{iso}\, \mathbf{F}_{iso}^{T}, \tag{6}$$

$$\mathbf{B}_{iso}^{v} = \mathbf{F}_{iso}^{v}\, \mathbf{F}_{iso}^{vT}, \tag{7}$$

$$\mathbf{B}_{iso}^{e} = \mathbf{F}_{iso}^{e}\, \mathbf{F}_{iso}^{eT}. \tag{8}$$

*2.2. Viscoelastic-viscoplastic damage model at finite deformation*

The Cauchy stress acting on the viscoelastic network is decomposed into equilibrium $\boldsymbol{\sigma}_{eq}$, non-equilibrium $\boldsymbol{\sigma}_{neq}$ and volumetric $\boldsymbol{\sigma}_{vol}$ terms. The equilibrium and non-equilibrium stress are given by a generalized neo-Hookean model

$$\boldsymbol{\sigma} = J^{-1}\left(\mu_{eq}\, \mathbf{B}_{iso}^{ve} + \mu_{neq}\, \mathbf{B}_{iso}^{e}\right), \tag{9}$$

and the volumetric part is defined by

$$\boldsymbol{\sigma}_{vol} = \frac{1}{2}k_v\left(J_m - \frac{1}{J_m}\right)\mathbf{1}, \tag{10}$$

resulting in an overall stress

$$\boldsymbol{\sigma}_{tot} = (\boldsymbol{\sigma} + \boldsymbol{\sigma}_{vol}), \tag{11}$$

which is formulated in the damaged state as follows:

$$\boldsymbol{\sigma}_{tot}^{d} = (1 - \mathrm{d})(\boldsymbol{\sigma} + \boldsymbol{\sigma}_{vol}), \tag{12}$$

where $\mathrm{d} \in [0, 1)$ is a scalar damage variable and its evolution obeys the following rule:

$$\dot{\mathrm{d}} = \mathrm{A}(1-\mathrm{d})\dot{\Lambda}_{chain}^{max}. \tag{13}$$

A is a material parameter calibrated using experimental data, $\Lambda_{chain} = \sqrt{\mathrm{tr}[\mathbf{B}_{iso}]/3}$, $\dot{\Lambda}_{chain}^{max}$ takes the following form:

$$\dot{\Lambda}_{chain}^{max} = \begin{cases} 0 & \Lambda_{chain} < \Lambda_{chain}^{max} \\ \dot{\Lambda}_{chain} & \Lambda_{chain} \geq \Lambda_{chain}^{max} \end{cases} \tag{14}$$

The shear moduli of the neo-Hookean stress contribution depend on the BNPs volume fraction $v_{np}$ and moisture content $w_w$ as follows:

$$\mu_{eq}(v_{np}, w_w) = \mathrm{X}(v_{np}, w_w)\, \mu_{eq}^{0}, \tag{15}$$

$$\mu_{neq}(v_{np}, w_w) = \mathrm{X}(v_{np}, w_w)\, \mu_{neq}^{0}, \tag{16}$$

while the volumetric bulk modulus $k_v$ is a constant. Nanoparticles play an important role in the mechanical behavior of the epoxy system. The BNPs are assumed to be rigid particles, which occupy a significant volume and serve as effective stiff fillers in the material. We apply a slightly modified Guth-Gold model [49] to obtain the effective stiffness of the nanoparticle-modified epoxy system. Although the quadratic form of the amplification factor X is adopted, we increase the gradient of the quadratic function, since the BNP reinforced epoxy is assumed to have a higher stiffness than the thermoplastic material used in [37]. Also, we add a moisture dependency on the effective stiffness of the material, resulting to the following amplification factor [3]

$$\mathrm{X} = \left(1 \;+\; 5v_{np} \;+\; 18v_{np}^2\right)\left(1 +\; 0.057w_w^2 \;-\; 9.5w_w\right), \tag{17}$$

where $v_{np}$ is the volume fraction of BNPs and $w_w$ represents the moisture content. The total velocity gradient of the viscoelastic network, $\mathbf{L}^{ve} = \dot{\mathbf{F}}^{ve}(\mathbf{F}^{ve})^{-1}$, can be decomposed into an elastic and a viscous component analogously to Eq. (5)

$$\mathbf{L}^{ve} = \mathbf{L}^e \;+\; \mathbf{F}^e \mathbf{L}^e \mathbf{F}^{e-1} \;=\; \mathbf{L}^e \;+\; \tilde{\mathbf{L}}^v, \tag{18}$$

and

$$\tilde{\mathbf{L}}^v = \dot{\mathbf{F}}^v \mathbf{F}^{v-1} \;=\; \tilde{\mathbf{D}}^v \;+\; \tilde{\mathbf{W}}^v. \tag{19}$$

Here, $\tilde{\mathbf{D}}^v$ represents the rate of the viscous deformation and $\mathbf{W}^v$ is a skew-symmetric tensor representing the rate of stretching and spin, respectively. We make the intermediate state unique by prescribing $\tilde{\mathbf{W}}^v = 0$. The rate of the viscoelastic flow is constitutively described by

$$\tilde{\mathbf{D}}^v = \; \frac{\dot{\varepsilon}^v}{\tau_{neq}} \; \mathrm{dev}\left[\boldsymbol{\sigma}'_{neq}\right] \tag{20}$$

where $\tau_{neq} = \|\mathrm{dev}[\boldsymbol{\sigma}_{neq}]\|_F$ represents the Frobenius norm of the driving stress, $\dot{\varepsilon}^v$ is the viscous flow and $\boldsymbol{\sigma}'_{neq} = \mathbf{R}_e^T \boldsymbol{\sigma}_{neq} \mathbf{R}_e$ represents the stress acting on the viscous component in its relaxed configuration. The viscous flow is defined by the Argon model:

$$\dot{\varepsilon}_v = \; \dot{\varepsilon}_0 \; \exp\left[\frac{\Delta \mathrm{H}}{k_b \mathrm{T}} \; \left(\left(\frac{\tau_{neq}}{\tau_0}\right)^m - 1\right)\right], \tag{21}$$

where $k_b$, $\dot{\varepsilon}_0$, $\Delta \mathrm{H}$ and $\tau_0$ are the Boltzmann constant, a pre-exponential factor, the activation energy and the athermal yield stress. Recent models [39, 50] have shown a better agreement with experimental data by using the exponential factor $m$ as a material parameter and therefore reconcile the moisture- or temperature dependency of the stiffness with the softening of the viscoelastic flow. We also modify the athermal yield stress of the argon model and propose a nonlinear behavior of the athermal yield stress driven by the local chain stretch $\Lambda_{chain}$, in contrast to the linear modification proposed by [37]. In our case, we use a sigmoid function for the athermal yield stress modification as follows:

$$\tau_0 = y_0 + \frac{a_s}{1 + \exp\left(-\frac{\left(\dot{\Lambda}_{chain}^{max} - x_0\right)}{b_s}\right)}. \tag{22}$$

The presented modification leads to an increase of the viscoelasticity with increasing stretch and damage. This leads to an increasing hysteresis after each cycle, which is also observed in different experimental data [43, 51]. Especially in the first few cycles an increasing hysteresis converging to a constant value after a certain point, is observed. Our model captures the initial exponential increase before reaching a plateau after a certain stretch. This way we ensure an inflationary increase of the viscoelastic behavior in the beginning of the stretch while keeping the athermal yield stress at the deformed configuration upon removal of applied stretch, thus assuming that the athermal yield stress change is taken to be permanent. The parameters $y_0$, $x_0$, $a_s$ and $b_s$ are material parameters which are calibrated using experimental data. The implemented sigmoid function is presented in Fig. 2.

In summary, the time derivative of $\dot{\mathbf{F}}^v$ can be derived from Eq. (20) and Eq. (19) as follows:

$$\dot{\mathbf{F}}^v = \mathbf{F}^{e-1}\dot{\varepsilon}^v \frac{\text{dev}\left[\boldsymbol{\sigma}'_{neq}\right]}{\tau_{neq}} \mathbf{F}^{ve}. \tag{23}$$

Similarly, the total velocity gradient of the overall network, $\mathbf{L} = \dot{\mathbf{F}}(\mathbf{F})^{-1}$, can be expanded

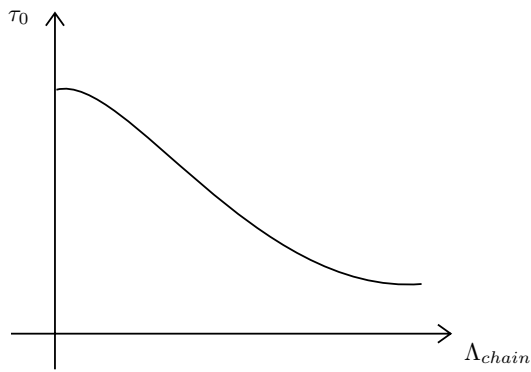

**Figure 2:** Proposed sigmoid-function for the behavior of the athermal yield stress $\tau_0$ driven by the local chain stretch $\Lambda_{chain}$.

to the following:

$$\mathbf{L} = \mathbf{L}^{ve} + \mathbf{F}^{ve}\mathbf{L}^{vp}\mathbf{F}^{ve-1} = \mathbf{L}^{ve} + \tilde{\mathbf{L}}^{vp}. \tag{24}$$

Again, we consider the viscoplastic velocity gradient to be additively decomposed into the symmetric rate of stretching and the skew-symmetric rate of spinning:

$$\tilde{\mathbf{L}}^{vp} = \dot{\mathbf{F}}^{vp}\mathbf{F}^{vp-1} = \tilde{\mathbf{D}}^{vp} + \tilde{\mathbf{W}}^{vp}, \tag{25}$$

and we take $\tilde{\mathbf{W}}^{vp} = 0$ again leading to:

$$\tilde{\mathbf{D}}^{vp} = \frac{\dot{\varepsilon}^{vp}}{\tau_{tot}} \operatorname{dev}\left[\boldsymbol{\sigma}'_{tot}\right], \tag{26}$$

where $\operatorname{dev}\left[\boldsymbol{\sigma}'_{tot}\right]$ is the total deviatoric stress in its relaxed configuration and $\tau_{tot}$ is the Frobenius norm of the total stress. To characterize the viscoplastic flow $\dot{\varepsilon}^{vp}$, we implement a simple phenomenological representation, similar to [52], as follows:

$$\dot{\varepsilon}^{vp} = \begin{cases} 0 & \tau_{tot} < \sigma_0 \\ a(\epsilon - \epsilon_0)^b \dot{\epsilon} & \tau_{tot} \geq \sigma_0 \end{cases}, \tag{27}$$

where a,b and $\sigma_0$ are material parameters. $\epsilon_0$ is the stress at which the viscoplastic flow is activated, represented by the Frobenius norm of the Green strain tensor $\parallel \mathbf{E} \parallel_F$, which is derived from the deformation gradient:

$$\mathbf{E} = \frac{1}{2}(\mathbf{F}^T\mathbf{F} - \mathbf{I}), \tag{28}$$

and $\dot{\epsilon}$ is the strain rate of the effective strain $\parallel \mathbf{E} \parallel_F$, thus introducing a simple strain-rate dependency of the viscoplastic flow. Analogous to Eq. (23), the time derivative of the viscoplastic deformation gradient is given by

$$\dot{\mathbf{F}}^{vp} = \mathbf{F}^{ve-1}\dot{\varepsilon}^{vp}\frac{\operatorname{dev}\left[\boldsymbol{\sigma}'_{tot}\right]}{\tau_{tot}}\mathbf{F}_{iso}, \tag{29}$$

characterizing the rate kinematics of the viscoplastic flow. We obtain the viscous and viscoplastic deformation gradients at the end of a time increment using the Euler backward time integration. The step-by-step procedure is presented in Section 4.

## 3. Deep learning-based constitutive model

To replace the complex viscoelastic-viscoplastic model with a deep learning-based model, the following framework is implemented:

1. Calibrate and validate the nonlinear viscoelastic-viscoplastic damage model using experimental data.
2. Use the following nonlinear mapping of the input sequence to the corresponding output sequence:

$$\begin{bmatrix} \vec{\mathbf{B}} \\ \Delta t \\ w_w \\ v_{np} \end{bmatrix} \rightarrow \begin{bmatrix} \vec{\sigma}_{tot} \end{bmatrix}, \tag{30}$$

$\vec{\mathbf{B}}$ represents the upper triangular components of the overall left Cauchy-Green deformation tensor, $\Delta t$ is the timestep, and $\vec{\sigma}_{tot}$ is the stress vector of the undamaged material containing the upper triangular components again. This ensures fewer inputs and outputs while capturing two physics-informed principles: Local balance of angular momentum and preserving the stress-free undeformed configuration [53]. The proposed mapping scheme leads to 9 input values mapped to 6 output values. We include $\Delta t, w_w$, and $v_{np}$ to ensure the effect of rate, moisture, and BNP content on the nanocomposites' behavior will be captured. Here, the stress tensor of the undamaged material is calculated after the mapping scheme to obtain more flexibility regarding the damage model.
3. Generate specific data by adapting a perturbation method to capture any complex loading condition and accurately calculate an approximation of the tangent modulus tensor $\hat{\mathbb{C}}$.
4. Train and validate an LSTM-enhanced deep network to learn the constitutive model.
5. Implement the proposed DL model in the FE analysis.

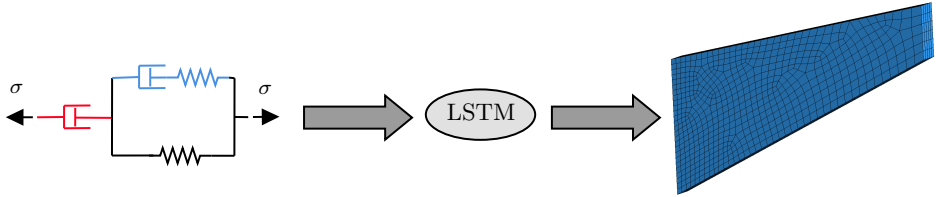

**Figure 3:** Proposed scheme for the finite element implementation of the intelligent constitutive model. Starting from the constitutive model, a DL model is developed and integrated into the finite element analysis.

In the following subsections, the architecture of the DL model and the training framework are presented.

*3.1. Long-short term memory network*

In deep learning, dense feed-forward neural networks are the basic building block of deep networks and can represent the nonlinear mapping of Eq. (30) from inputs **x** to predictions **t** with a number of consecutive layers L and trainable parameters $\boldsymbol{\omega}$ as

$$\mathcal{F}_{nn}(\boldsymbol{\omega}): \mathbf{x} \rightarrow \mathbf{t}, \tag{31}$$

where the trainable parameters $\boldsymbol{\omega}$ include weights **W** and biases **b** of each Layer. For each layer, a linear transformation of the inputs **x** is applied before a non-linear activation is enforced to obtain the activation of each layer $\mathbf{a}^l$. This can be expressed as

$$\mathbf{a}^l = \Phi^l\left(\mathbf{W}^l \cdot \mathbf{a}^{l-1} + \mathbf{b}^l\right), \, l = 1, 2 \, ... \, L. \tag{32}$$

Here, $\Phi$ is the nonlinear activation function and the first activation $\mathbf{a}^0$ corresponds to the input vector while the last activation $\mathbf{a}^L$ represents the output of the deep network.

While feed-forward deep networks have been proven to yield good results for the nonlinear behavior of constitutive models [11, 54–56], they are not able to predict sequential data $\mathbf{x}^t$ that may be sorted according to real-time and collected into a larger sequence $\mathbf{x} = [\mathbf{x}^1...\mathbf{x}^T]$. Hence, we implement a multilayered LSTM deep network consisting of several memory cells and gates for remembering and forgetting information in each sequence by minimizing a loss between the target **t** and the prediction $\mathbf{t}^*$. The input sequence in our case yields $\mathbf{x} = ([\vec{\mathbf{B}}, \Delta \mathrm{t}, w_w, v_{np}]^1, \, ... \, , [\vec{\mathbf{B}}, \Delta \mathrm{t}, w_w, v_{np}]^T)$, which is mapped to a stress vector $\mathbf{t} = ([\vec{\boldsymbol{\sigma}}_{tot}]^1, \, ... \, , [\vec{\boldsymbol{\sigma}}_{tot}]^T)$.

The architecture of a single LSTM cell is presented in Fig. 4. In the illustration, **h** and **c** denote the hidden and cell states at timestep $i-1$ and $i$, respectively. The hidden state is an encoding of the most recent timestep and can be processed at any point to obtain meaningful data. The cell state acts as a global memory of the LSTM network over all timesteps, allowing the LSTM cell to have information on the history of each sequence. The learnable parameters $\boldsymbol{\omega}$ of each component presented in Fig. 4 of the LSTM cell are the input weights **W**, the recurrent weights **R**, and the bias **b**. These matrices are concatenations of the input weights, the recurrent weights, and the bias of each gate as follows:

$$\mathbf{W} = \begin{bmatrix} \mathbf{W}_i \\ \mathbf{W}_f \\ \mathbf{W}_g \\ \mathbf{W}_o \end{bmatrix}, \ \mathbf{R} = \begin{bmatrix} \mathbf{R}_i \\ \mathbf{R}_f \\ \mathbf{R}_g \\ \mathbf{R}_o \end{bmatrix}, \ \mathbf{b} = \begin{bmatrix} \mathbf{b}_i \\ \mathbf{b}_f \\ \mathbf{b}_g \\ \mathbf{b}_o \end{bmatrix}, \tag{33}$$

where $i, f, g$ and $o$ denote the input gate, forget gate, cell candidate, and output gate, respectively, as presented in Fig. 4.

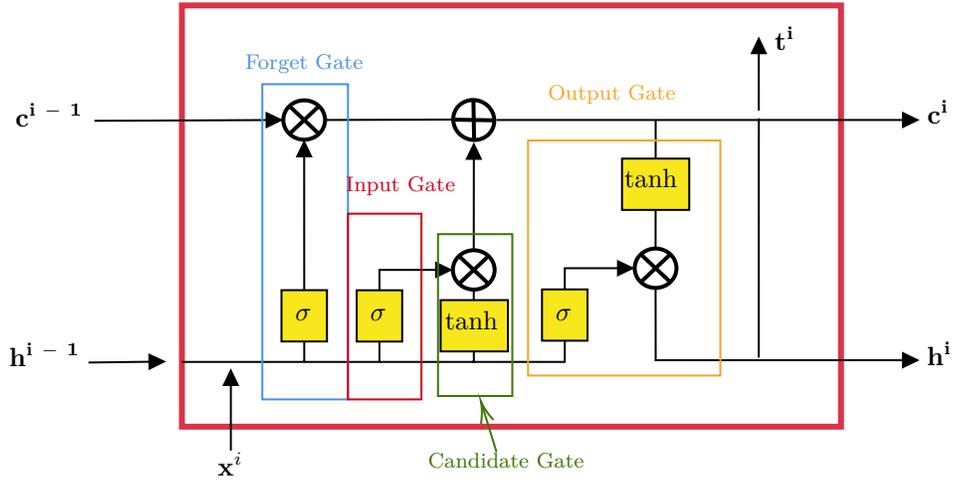

**Figure 4:** A single LSTM cell consisting of multiply connected layers. $\boldsymbol{\sigma}$ represents the sigmoid function, *tanh* is the hyperbolic tangent function. The forget gate, input gate and output gate do control the data flow into the cell. The candidate gate presents a possible candidate for the cell state.

The cell state $\boldsymbol{c}^i$ at timestep $i$ is given by

$$\mathbf{c}^i = \mathbf{f}^i \odot \mathbf{c}^{i-1} + \mathbf{i}^i \odot \mathbf{g}^i. \tag{34}$$

The operator $\odot$ corresponds to Hadamard product or element-wise product. The hidden state $\mathbf{h}^i$ can be evaluated as follows:

$$\mathbf{h}^i = \mathbf{o}^i \odot \tanh\left(\mathbf{c}^i\right), \tag{35}$$

where the components of each gate are described by the following equations:

$$\mathbf{i}^i = \sigma\left(\mathbf{W}_i \mathbf{x^i} + \mathbf{R}_i \mathbf{h}^{i-1} + \mathbf{b}_i\right), \tag{36}$$

$$\mathbf{f}^i = \sigma\left(\mathbf{W}_f \mathbf{x}^i + \mathbf{R}_f \mathbf{h}^{i-1} + \mathbf{b}_f\right), \tag{37}$$

$$\mathbf{g}^i = \tanh\left(\mathbf{W}_g \mathbf{x}^i + \mathbf{R}_g \mathbf{h}^{i-1} + \mathbf{b}_g\right), \tag{38}$$

$$\mathbf{o}^i = \sigma\left(\mathbf{W}_o \mathbf{x}^i + \mathbf{R}_o \mathbf{h}^{i-1} + \mathbf{b}_o\right). \tag{39}$$

In the equations above, $\mathbf{i}^i$, $\mathbf{f}^i$, $\mathbf{g}^i$ and $\mathbf{o}^i$ are respectively the components of the input gate, forget gate, cell candidate and output gate, and $\boldsymbol{\sigma}$ represents the sigmoid function. The element-wise product allows each gate to control the data flow into the cell state $\mathbf{c}^i$ by considering the history of the sequence. As we can see in Eq. (34), we use the previous cell state and the current cell candidate to obtain the current cell state $\boldsymbol{c}^i$, which then is used as an input to obtain the final output vector $\mathbf{t}^i$ at each timestep as follows:

$$\mathbf{t}^i = \mathbf{o}^i \odot \tanh\left(\mathbf{c}^i\right). \tag{40}$$

To predict the stress tensor using the described Deep Network, we employ two LSTM layers and connect them to a dense forward layer. Therefore, we need to save two hidden states and cell states for each LSTM cell as depicted in Fig. 5. The stacked LSMT units are responsible for processing the input vector $\mathbf{x}^i$ and the state variables through the gates, therefore updating the state of the LSTM units and representing an encoded representation of the current state. This ensures that the required history path dependency is included. To transform the encoded representation of the current state from the LSTM units, we apply a dense forward layer that approximates the stress vector $\boldsymbol{\sigma}^i_{tot}$ from the LSTM output $\mathbf{t}^i$.

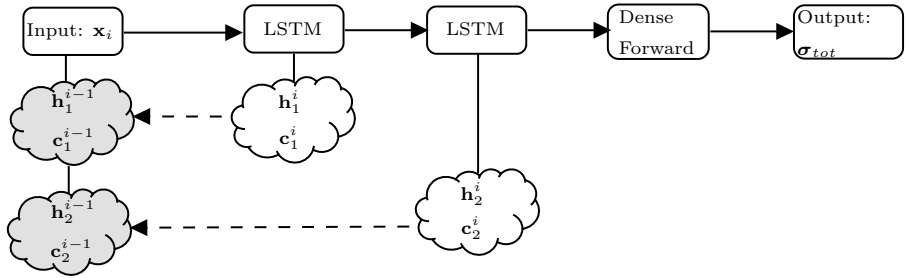

**Figure 5:** Overall structure of the LSTM enhanced intelligent constitutive model including two LSTM layers. The input layer includes our input vector $\mathbf{x}_i$ at step $i$ and the state variables $\mathbf{h}_1^{i-1}$, $\mathbf{c}_1^{i-1}$, $\mathbf{h}_2^{i-1}$ and $\mathbf{c}_2^{i-1}$ used in both LSTM layers. The output vector $\boldsymbol{\sigma}_{tot}$ is predicted by the dense forward layer.

We follow a supervised approach to train our parameters $\boldsymbol{\omega}$ of our deep network. The subsection below presents a space-filling sampling approach to generate training data.

*3.2. Data generation*

To capture the path-dependent behavior of our constitutive model, we presented a mapping scheme including moisture, strain rate, and BNPs volume fraction dependency. Consequently, generating the database for the supervised learning of the LSTM-enhanced Deep Network as presented in Fig. 5 is essential for the learning process. Different approaches are presented in the literature to generate training data [23, 27, 53, 57–60]. Here, we implement a space-filling procedure to make sure our DL-model can be trained sufficiently to predict the stress $\boldsymbol{\sigma}_{tot}$ of the viscoelastic-viscoplastic constitutive model at any possible three-dimensional state.

The driving force for the generation of loading paths in our finite deformation model is the deformation gradient **F**. Each $F_{ij}(i, j = 1, ..., 3)$ component leads to a different loading scenario. Therefore, we generate data using the deformation gradient in a spatiotemporal space. Firstly, we constrain the components of the deformation gradient as follows:

$$F_{ij} \in \begin{cases} [0.9 \quad 1.1], & \text{when i } = \text{ j} \\ [-0.05 \quad 0.05], & \text{when i } \neq \text{ j} \end{cases}. \tag{41}$$

In this study, the sampling process starts from an undeformed configuration in which the diagonal elements of the deformation gradient are set to 1.0, and the upper/lower triangular components are set to 0.0. This ensures a bounded spatial space within a realistic viscoelastic-viscoplastic regime for the nanocomposites. Accordingly, the DL model is expected to learn and predict the correct output accurately inside this domain. To cover the nine-dimensional spatial space with sufficient random points, quasi-random numbers are produced using the Halton sequence generation algorithm [61], leading to uniform samples within the space and a better sampling of the region. As presented in Fig. 6, the resulting points in the spatial space adequately span the bounded domain compared with the pseudorandom algorithm used in most algorithms.

To generate loading paths for training the DL model, we first produce the uniform data points for each of the nine components of the total deformation gradient. It allows capturing loading scenarios related to uniaxial tension or compression, triaxial loading, biaxial loading, and pure or simple shear loading. We utilize the uniform data points for each component of the deformation gradient to generate 5% of the training data. Secondly, we implement an algorithm to randomly visit the quasi-random data points in the nine-dimensional spatial

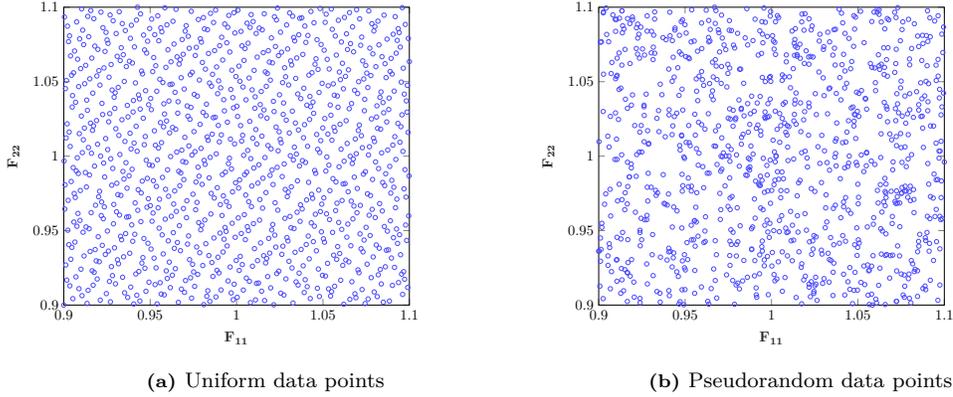

**(a)** Uniform data points

**(b)** Pseudorandom data points

**Figure 6:** Example of the generated data points in a two dimensional space using the Halton sequence algorithm (a) and the pseudorandom MATLAB algorithm (b).

space of the deformation gradient components. Therefore, loading paths to capture complex loading scenarios are created as shown, for example, in Fig. 7 for the diagonal components of the deformation gradient. We utilize the quasi-random data points in the nine-dimensional space to generate 95% of the training data. This process ensures that both simple and complex loading scenarios are included in the training data.

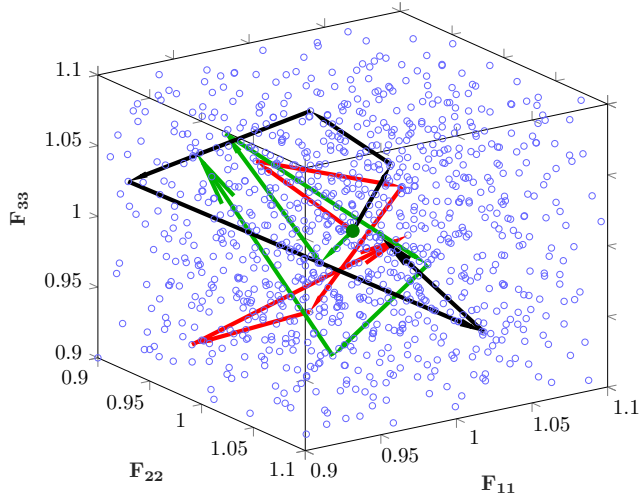

**Figure 7:** Example of three generated loading paths in the uniformly distributed space for the diagonal components $F_{ii}$ of the deformation gradient. All paths start from the undeformed configuration (i.e., $\mathbf{F} = \mathbf{I}$).

The created sequence then serves as an input to integrate the constitutive model using the Euler backward algorithm and create the training sequence as presented in Eq. (30).

The loading paths are created using different time and deformation increments within $\Delta \mathrm{F} \in \left[10^{-6}, 10^{-4}\right]$ and $\Delta \mathrm{t} \in [0.05, 5]$s. This ensures a realistic time and deformation step within FE simulations and allows us to constrain the strain rate within $\dot{\varepsilon} \in \left[10^{-5}, 10^{-3}\right]$ 1/s. Table 1 summarizes the step-by-step algorithm for generating training data in the nine-dimensional spatiotemporal space.

**Table 1:** Summary of the step-by-step algorithm for generation of a single training sequence.

1. Generate uniform data for the nine components.
   of the deformation gradient in a range defined by Eq.(41).
2. Define P as the number of points to be visited.
3. Generate loading path within the 9-dimensional space
   as exemplary presented in Fig. 8 (for the 3-dimensional space)
   for moisture content $w_w$ and BNPs volume fraction $v_{np}$.
4. Calculate a representative strain rate as: $\dot{\epsilon} = \| \mathbf{E} \|_F \ / \ \Delta \mathrm{t}$,
   where $\mathbf{E}$ is presented in Eq. (28).
5. If $\dot{\epsilon} > 1 \cdot 10^{-5}$ $and$ $\dot{\epsilon} < 1 \cdot 10^{-3}$ GOTO step 6 else GOTO step 3.
6. Add a perturbation step to the generated loading path according to algorithm 1.
7. Integrate the constitutive model and obtain $\boldsymbol{\sigma}_{tot}$.
8. Create the input sequence $\mathbf{x}$ and the output sequence $\boldsymbol{\sigma}_{tot}$ for training.

This process is repeated for three different points to be visited P: $\mathrm{P} = 1, \mathrm{P} = 3$ and $\mathrm{P} = 6$. P represents the number of points visited within a loading path in the space, thus resulting in different loading-unloading scenarios in tension and compression and capturing complex deformation paths. The overall generated data for the supervised learning contains $\mathrm{T} = 52000$ sequences, including 10% for validation. For better performance of the DL model, we normalize our input data by calculating the per-feature mean and standard deviation of all the sequences. Then, we subtract the mean value and divide each training observation by the standard deviation. The training is done on two Tesla V100 GPUs, with each 20 CPU cores.

For FE analysis, we need to accurately predict the stress tensor $\boldsymbol{\sigma}_{tot}$ for the constitutive model and the perturbation method, as presented in the next section. For this, each generated loading path is modified by adding a random perturbation at the end of each sequence as presented in algorithm 1 and illustrated exemplarily in Fig. 9. Accordingly to the algorithm, the components to be perturbed are chosen randomly. Therefore, in this specific case, no perturbation associated with the component $\mathrm{F}_{22}$ is employed. This allows the deep network to randomly learn a sudden change of each deformation gradient component by a specific perturbation step, which is also the case for the perturbation method. It is noteworthy that the tangent modulus $\hat{\mathbb{C}}$ can also be computed from the automatic differentiation of the DL model [15, 62, 63]. However, due to the scarcity of automatic differentiation libraries in finite element software packages and the higher computational efficiency [23], we propose a data generation scheme to predict the tangent modulus using the perturbation method accurately.

---
**Algorithm 1** Perturbation of the loading path in order to learn the approximation of $\hat{\mathbb{C}}$
---
    **Input**: Total loading path: F with length n
    **Output**: Perturbed last step included in F
1: Perturbation parameter: $\alpha = 1 \cdot 10^{-4}$
2: GOTO last step of loading path: $F_{ij}(n)$
3: Choose random components $i_p$ and $j_p$ of $F_{ij}(n)$ to be perturbed
4: **for** i = 1:3 **do**
5:     **for** j = 1:3 **do**
6:         **if** $i == i_p$ and $j == j_p$ **then**
7:             Perturbate component: $\tilde{F}_{ij} = F_{ij}(n) + \alpha$
8:             Add component to F: $F_{ij}(n+1) = \tilde{F}_{ij}$
9:         **else**                                                                     ▷ Needed to keep sequence length constant
10:             Add unchanged component to F: $F_{ij}(n+1) = F_{ij}(n)$
11:         **end if**
12:     **end for**
13: **end for**

---

With the technique used, a training framework to obtain data-driven surrogate models at finite deformation can be developed. The training data is able to inherently capture four basic concepts of our constitutive model:

- Viscoelasticity and viscoplasticity in a three-dimensional space.
- Strain rate dependency within the range of $1 \cdot 10^{-5}$ and $1 \cdot 10^{-3}$.
- Moisture and BNPs volume fraction dependency.
- Approximation of the tangent modulus tensor using the perturbation method.

*3.3. Hyperparameters*

The hyperparameters of the Deep Network include the number of LSTM cell layers, the activation function for the output layer, the batch size, the number of units in an LSTM layer,

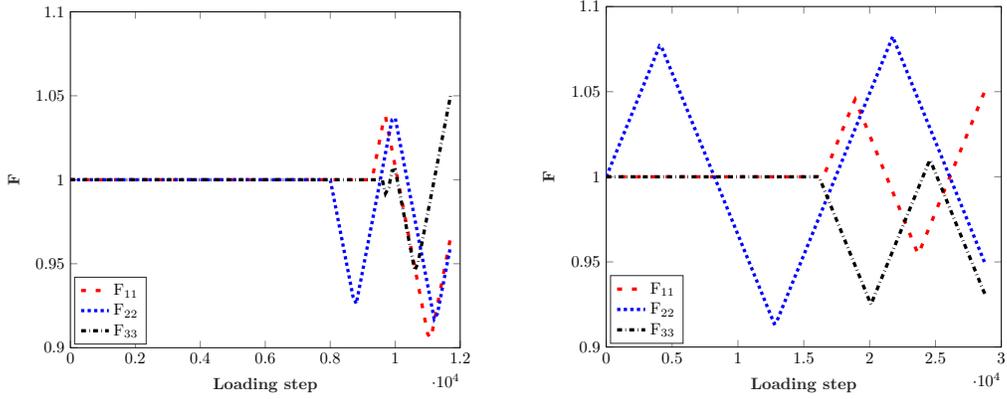

(a) Sequence corresponding to the red path in Fig. 7.  (b) Sequence corresponding to the green path in Fig. 7.

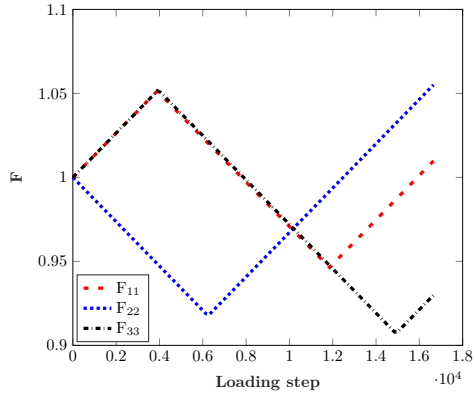

(c) Sequence corresponding to the black path in Fig. 7.

**Figure 8:** The created loading paths for the diagonal terms of the deformation gradient generated from the visited points as displayed in Fig. 7. Each one represents an unique sequence considered in the training data.

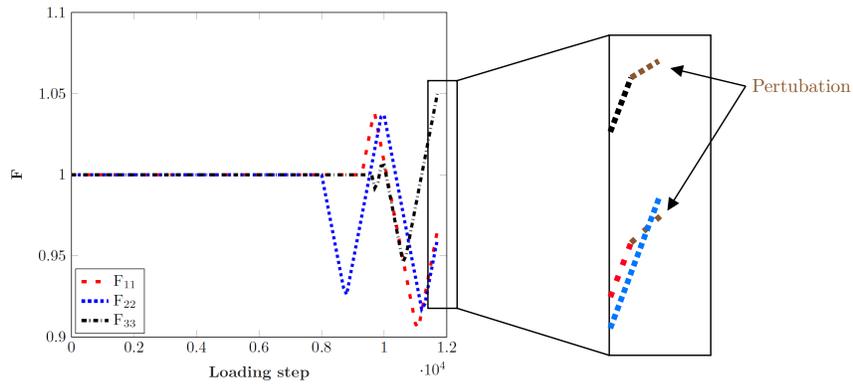

**Figure 9:** Exemplary illustration of the algorithm 1 used on the loading path presented in Fig. 8a.

and the number of epochs for training. The architecture of the Deep Network consists of two layers of LSTM units connected to a dense forward layer. We employ the sigmoid activation function for the output layer and train for 300 epochs to reach a final training state.

In the next section we present a finite element formulation implemented to solve the linear equilibrium equation.

## 4. Finite-element formulation

The Euler-Langrange equations for the strong form of the boundary value problem in referential form can be written as

$$\nabla_x \cdot \mathbf{P} + \mathbf{B} = 0 \text{ in } \Omega, \tag{42}$$

$$\mathbf{P} \cdot \mathbf{N} = \overline{\mathbf{T}} \text{ on } \Gamma_T, \tag{43}$$

$$\mathbf{u} = \mathbf{u}_d \text{ on } \Gamma_d, \tag{44}$$

where $\mathbf{P}$ is the first Piola-Kirchhoff stress, B is the vector of body forces in referential form on the body $\Omega$, N is the outward unit normal vector on the boundary $\Gamma_T$, $\overline{\mathbf{T}}$ is the traction force and $\mathbf{u}_d$ represents the prescribed displacements at the boundary $\Gamma_d$. To obtain the weak forms of Eq. (42), a multiplication of the residual by a weighting function $\eta_u$ and by integrating the residual over the whole domain is fulfilled. Using the Gauss divergence theorem leads to the following equations:

$$\int_{\Omega_0} \mathbf{P} \cdot \nabla_x \eta_u \, d\Omega_0 - \int_{\Omega_0} \rho_0 \mathbf{B} \cdot \eta_u \, d\Omega_0 - \int_{\Gamma_0} \overline{\mathbf{T}} \cdot \eta_u \, d\Gamma_0 = 0. \tag{45}$$

The equation can then be expressed in terms of external and internal nodal forces as:

$$\mathbf{r}^u = f_{int}^u - f_{ext}^u = 0, \tag{46}$$

where

$$f_{int}^u = \int_{\Omega_0} \mathbf{P} \cdot \nabla_x \eta_u \, d\Omega_0, \tag{47}$$

and

$$f_{ext}^u = \int_{\Omega_0} \rho_0 \mathbf{B} \cdot \eta_u \, d\Omega_0 + \int_{\Gamma_0} \overline{\mathbf{T}} \cdot \eta_u \, d\Gamma_0. \tag{48}$$

By linearizing Eq. (46) at iteration $i+1$ with respect to the previous iteration $i$ and assuming dead loads:

$$\mathbf{r}_{i+1}^u = \mathbf{r}_i^u + \Delta \mathbf{r}^u = 0, \tag{49}$$

where

$$\Delta \mathrm{r}^u = \mathrm{D}_u \mathrm{r}_i^u \cdot \Delta \mathrm{u}, \tag{50}$$

and

$$\mathrm{D}_u \mathrm{r}_i^u \cdot \Delta \mathrm{u} = \int_{\Omega_0} \mathrm{D}_u \mathbf{P} \cdot \Delta \mathrm{u} \cdot \nabla_x \eta_u \ \mathrm{d}\Omega_0, \tag{51}$$

which represents the directional derivative of the operator $\mathrm{r}^u$ in the direction of $\Delta \mathrm{u}$. The linearization of the first Piola-Kirchhoff stress tensor leads to $\mathbf{P} = \mathbf{FS}$ and the linearization of the second Piola-Kirchhoff stress tensor can be based on the following equation:

$$\mathrm{D}\mathbf{S} \cdot \Delta \mathrm{u} = \mathbb{C}\Delta \mathbf{E}, \tag{52}$$

where the last term is the linearization of the Green-Lagrange strain tensor $\mathbf{E}$ and $\mathbb{C}$ is the elasticity tensor which is referred to the initial configuration. The linearization of the weak form is completed as follows:

$$\begin{aligned}
\mathrm{D}_u \mathrm{r}_i^u \cdot \Delta \mathrm{u} &= \int_{\Omega_0} \left( \nabla_x \Delta \mathrm{u}\ \mathbf{S} + \mathbf{F}\mathrm{D}_u \mathbf{S}\Delta \mathrm{u} \right) \cdot \nabla_x \eta_u \ \mathrm{d}\Omega_0 \\
&= \int_{\Omega_0} \left( \nabla_x \Delta \mathrm{u}\ \mathbf{S} + \mathbf{F}\mathbb{C}\Delta \mathbf{E} \right) \cdot \nabla_x \eta_u \ \mathrm{d}\Omega_0.
\end{aligned} \tag{53}$$

The linearization of the weak form in the spatial configuration can be obtained now by a *push forward* in $\nabla_x^s \Delta \mathrm{u}$ of the linearization in Eq. (53) to the known current configuration, which leads to the updated Lagrange formulation:

$$\mathrm{D}_u \mathrm{r}_i^u \cdot \Delta \mathrm{u} = \int_{\Omega_t} \left( \nabla_x \Delta \mathrm{u}\ \boldsymbol{\sigma} \cdot \nabla_x \eta_u + \nabla_x^s \eta_u \cdot \hat{\mathbb{C}} \nabla_x^s \eta_u \Delta \mathrm{u} \right) \ \mathrm{d}\Omega_t, \tag{54}$$

where $\boldsymbol{\sigma}$ is now the Cauchy stress. The Eq. (54) is referred to the current configuration, which leads to the following definition of the constitutive tensor:

$$\hat{\mathbb{C}} = \frac{1}{J}\mathbb{C}. \tag{55}$$

For a more specific and detailed discussion, the reader is referred to [64]. Employing the Bubnov-Galerkin method, the displacement and the corresponding weight functions are discretized in each element by

$$\mathbf{u}^h = \mathbf{N}\mathbf{u}, \ \boldsymbol{\eta}_u^h = \mathbf{N}\boldsymbol{\eta}_u, \ \nabla \mathbf{u}^h = \mathbf{B}\mathbf{u}, \tag{56}$$

where the shape function matrix $\mathbf{N}$ interpolates the nodal values $\mathbf{u}$, and $\mathbf{B}$ is the gradient operator for the displacements.

Substituting the relations into the weak formulation of the governing equations yields

$$\mathbf{K}_i \Delta \mathbf{U}_{i+1} = \mathbf{f}_{ext} - \mathbf{f}_{int,i}, \tag{57}$$

where

$$\mathbf{K}_i = \int_{\Omega_t} \mathbf{B}^T \hat{\mathbb{C}} \mathbf{B} \, d\Omega_t \, + \, \int_{\Omega_t} \mathbf{B}^T \boldsymbol{\sigma} \mathbf{B} \, d\Omega_t, \tag{58}$$

represents the linear and nonlinear stiffness matrices, respectively. We solve the presented linearization in Eq. (57) by using the Newton-Raphson iteration until the relative $L_2$-norm of the residual is less than a tolerance of $10^{-4}$. The step-by-step procedure is summarized in Table 2.

As mentioned above, we need the tangent modulus tensor $\hat{\mathbb{C}} = \dfrac{\partial \boldsymbol{\sigma}}{\partial \boldsymbol{\varepsilon}}$ to integrate our material model into a finite element framework. Since a closed form is not a straightforward task, we adapt the approach proposed by Sun et al. [65] to estimate the tangent moduli for the Jaumann rate of the Kirchhoff stress, $\boldsymbol{\tau} = J\boldsymbol{\sigma}$. We obtain the tangent modulus by perturbing the (i,j) components of the deformation gradient $\mathbf{F}_{ij}$. The choice of (ij) is chosen to be (11),(22),(33),(12),(13), and (23), leading to the following equation:

$$\mathbb{C} \approx \frac{1}{\alpha} \left( \boldsymbol{\tau}\left(\hat{\mathbf{F}}^{ij}\right) - \boldsymbol{\tau}(\mathbf{F}) \right), \tag{59}$$

where $\hat{\mathbf{F}}^{ij} = \mathbf{F} + \Delta \mathbf{F}_{ij}$ is the perturbed deformation gradient and $\alpha$ is the perturbation parameter. The final tangent modulus tensor is then obtained using Eq. (55). The reader is referred to [65] for a detailed discussion on the numerical approximation of the tangent modulus tensor.

As described above, we implement the Euler backward method to integrate the state variables of the constitutive model. The Euler backward method is an iterative scheme. Accordingly, the computational cost increases with highly nonlinear behavior captured by our constitutive model. Also, besides integrating the state variables described in steps 5 and 6, we need to integrate our constitutive model another six times by perturbing the deformation gradient to obtain the tangent modulus tensor as shown in Eq. (59).

## 5. Results and discussion

In the following section, the constitutive model is first calibrated using experimental data. Due to the chosen geometry of the specimen used within our experiments, we utilize FE

**Table 2:** Summary of the step-by-step algorithm for the integration of state variables.

1. Known values at time $t$:
   - Deformation gradient: ${}^t\mathbf{F}$, ${}^t\mathbf{F}_{iso}$, ${}^t\mathbf{F}_{iso}^e$ and ${}^t\mathbf{F}_{iso}^{ve}$.
   - State variables: ${}^t\mathbf{F}_{iso}^v, {}^t\mathbf{F}_{iso}^{vp}$.

2. Known values at time $t + \Delta t$:
   - Deformation gradient: ${}^{t+\Delta t}\mathbf{F} + \Delta t$ and ${}^{t+\Delta t}\mathbf{F}_{iso} + \Delta t$.

3. Calculate trial isochoric viscoelastic and elastic deformation gradient using Eq. (4) and Eq. (5).

4. Calculate $\boldsymbol{\sigma}_{neq}$ using Eq. (9) at $t + \Delta t$.

5. Update ${}^t\mathbf{F}_{iso}^v$ using the Euler backward method and Eq. (23) for $\dot{\mathbf{F}}_{iso}^v$ obtaining $\dot{\mathbf{F}}_{trial/iso}^v$ as follows:
   - Calculate the trial viscous flow rate $\dot{\varepsilon}_v$ using Eq. (21).
   - Calculate the trial viscous stretching $\tilde{\mathbf{D}}^v$ using Eq. (20).
   - Calculate the trial viscous state variable:
     $${}^{t+\Delta t}\mathbf{F}_{trial/iso}^v = \left({}^{t+\Delta t}\mathbf{F}_{trial/iso}^e\right)^{-1} {}^{t+\Delta t}\mathbf{F}_{trial/iso}^{ve} + \Delta t.$$
   - Update the state variable: ${}^{t+\Delta t}\mathbf{F}_{iso}^v = {}^t\mathbf{F}_{iso}^v + \Delta t \tilde{\mathbf{D}}^v \, {}^{t+\Delta t}\mathbf{F}_{trial/iso}^v$.

6. Update ${}^t\mathbf{F}_{iso}^{vp}$ using the Euler backward method and Eq. (29) for $\dot{\mathbf{F}}_{iso}^{vp}$ obtaining $\dot{\mathbf{F}}_{trial/iso}^{vp}$ as follows:
   - Calculate the trial viscoplastic flow rate $\dot{\varepsilon}_{vp}$ using Eq. (27).
   - Calculate the trial viscoplastic stretching $\tilde{\mathbf{D}}^{vp}$ using Eq. (26).
   - Calculate the trial viscoplastic state variable:
     $${}^{t+\Delta t}\mathbf{F}_{trial/iso}^{vp} = \left({}^{t+\Delta t}\mathbf{F}_{trial/iso}^{ve}\right)^{-1} {}^{t+\Delta t}\mathbf{F}_{iso} + \Delta t.$$
   - Update the state variable: ${}^{t+\Delta t}\mathbf{F}_{iso}^{vp} = {}^t\mathbf{F}_{iso}^{vp} + \Delta t \tilde{\mathbf{D}}^{vp} \, {}^{t+\Delta t}\mathbf{F}_{trial/iso}^{vp}$.

7. If $\|\,{}^{t+\Delta t}\mathbf{F}_{iso}^v - {}^{t+\Delta t}\mathbf{F}_{trial/iso}^v\,\|$ and $\|\,{}^{t+\Delta t}\mathbf{F}_{iso}^{vp} - {}^{t+\Delta t}\mathbf{F}_{trial/iso}^{vp}\,\| <$ tolerance than GOTO step 8 else GOTO step 3.

8. Calculate $\boldsymbol{\sigma}_{eq}$ using Eq. (9) at $t + \Delta t$.

9. Update the damage variable at $t + \Delta t$ according to Eq. (13).

10. Obtain the total Cauchy stress using Eq. (9) at $t + \Delta t$.

11. Store the state variables: ${}^{t+\Delta t}\mathbf{F}_{iso}^v, {}^{t+\Delta t}\mathbf{F}_{iso}^{vp}$.

12. Compute the tangent modulus $\hat{\mathbb{C}}$.

13. Solve the system of equations in Eq. (57) using the Newton-Raphson method.

analysis to obtain realistic material parameters. Next, we present the training setup, and the simulation results of the DL model are compared with those of the constitutive model. Finally, the DL model's capability in predicting the force-displacement behavior within the finite element framework is evaluated, and its computational efficiency is compared with the constitutive model.

*5.1. Experiments*

The specimens for the conditioning and mechanical tests are cut from the panels as presented in Fig. 10. Works from Poulain et al. [39] and [43] suggest that the necking in tension direction of epoxy systems is instead a structural instability than material property, consequently a notch is inserted to reduce the influence of material imperfections and necking on its yield. The specimens are conditioned at 60 °C and 85% relative humidity until the saturated state at a moisture concentration of 1.0% for the neat epoxy system and 1.2% for the BNPs reinforced epoxy is reached. The overall conditioning time was 115 days. Finally, mechanical loading-unloading tests are produced according to testing standard DIN EN ISO 527-2, using an extensometer to measure the elongation of the specimens and a load rate of 1 mm/min. We apply a total of six cycles by loading to a certain amplitude and unloading until the loading force reaches zero.

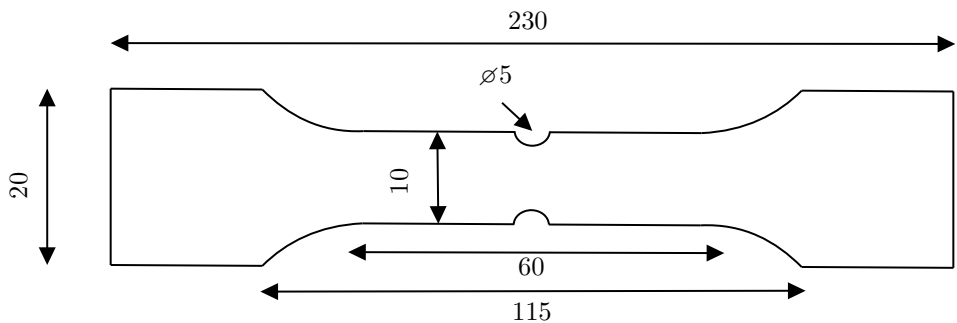

**Figure 10:** Planar dimensions of the specimen for conditioning and mechanical loading-unloading tests with a thickness of 2.3 mm. All dimensions are in millimeters.

*5.2. Calibration of the viscoelastic-viscoplastic damage model*

As presented above, the specimen used in our experiments includes stress concentration at the center of the sample to more accurately calibrate the damage parameter. Accordingly, the parameter identification of the constitutive model is conducted in two steps as follows:

1. The material parameters are initially pre-calibrated using the rheological model and experimental data. For this, the objective function, defined by the root mean square deviation between the experimentally measured and numerically predicted stress values, is minimized using a genetic algorithm. The population size is set at 200, and a maximum number of generations of 500 is used. To ensure that the optimum solution is obtained, the number of steps to determine whether the genetic algorithm is progressing is set to 500. The pre-calibration allows obtaining a precise upper and lower bound for the next parameter identification step.

2. Since the presented geometry of the specimens used in our experimental data cannot be considered within our one-dimensional rheological model, the final calibration of the material parameters is fulfilled using FE analysis. Also, the damage is concentrated at the center of the sample and can only be accurately calibrated using FE analysis. Therefore, this step re-optimizes the shear and volumetric bulk modulus of the equilibrium, the shear modulus of the non-equilibrium stress contribution, and the damage parameter using FE simulation results and experimental data. Here, an objective function is defined by the root mean square deviation between the experimentally measured and numerically predicted force values.

It is worth noting that although the viscous dashpot parameters can be identified using the experimental data, Unger et al. [42] performed a set of atomistic simulations and predicted the parameters of the Argon viscoelastic model. Accordingly, we adopt the parameter $\dot{\varepsilon}_0$ and the equivalent activation energy $\Delta H$. Furthermore, since FE simulations are computationally expensive, we calibrate the viscoplastic parameters manually by doing multiple sensitivity simulations to reduce the calibration time.

Considering the double symmetry of the specimen at mid-length, symmetric boundary conditions are applied in our finite element analysis to reduce the computational cost while keeping the full solution of the model. A three-dimensional simulation is performed to cover the transverse strain effect and loading conditions accordingly to the experiments are applied as presented in Fig. 11. The presented model is discretized with 16104 eight-noded hexahedral (Q8) elements. The mesh is refined at the right part of the model, where a stress concentration area is observed and 8 elements are implemented to discretize the thickness direction. We multiply the horizontal displacement at the loading point by two to measure the displacement. The final identified material parameters are listed on Table 3 and the corresponding force-displacement behavior is presented in Fig. 12.

The constitutive material model is validated using experimental data for the epoxy system with 0% BNPs at the saturated condition and the epoxy system including 10% BNPs at dry and saturated conditions. The prediction of the calibrated constitutive model for the epoxy system with 0% BNPs at saturated condition is presented in Fig. 13. The agreement between experimental data and constitutive model prediction in the figure confirms the predictive capability of the implemented viscoelastic-viscoplastic damage model on the mechanical behavior of the nanocomposite. Accurate results are also observed for the epoxy system containing 10% BNPs under saturated and dry conditions as presented in Fig. 14.

Although the proposed constitutive model can reasonably predict the highly nonlinear viscoelastic-viscoplastic behavior of the material, it has some shortcomings compared with the experimental data. The deviation from the experimental results may result from the formulation of the constitutive model and the unique set of material parameters. Especially in the first cycles at 0% BNPs volume fraction, the predicted stiffness and plastic strain at zero forces are overestimated. Nevertheless, the implemented amplification approach to include the influence of moisture and BNPs volume fraction leads to realistic results. It is visible from the numerical response of the 10% BNP/epoxy model at the saturated condition in Fig. 14, where both linear and nonlinear rate-dependent behaviors are appropriately captured.

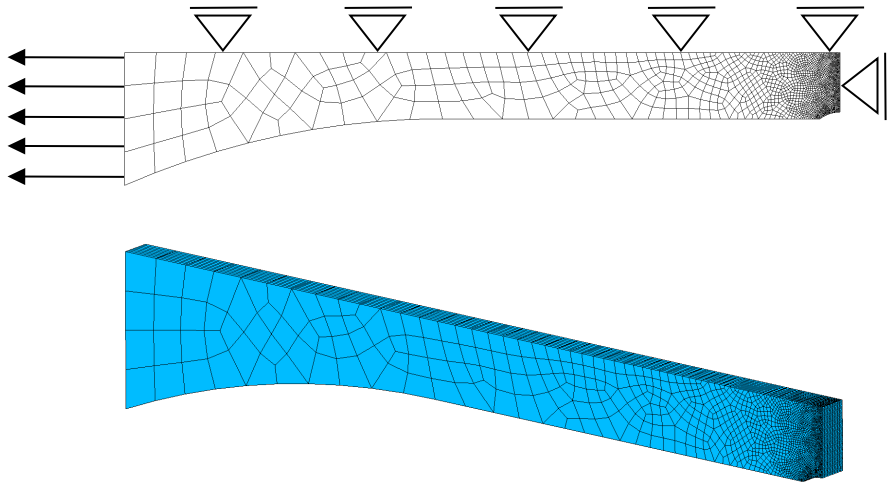

**Figure 11:** Loading and boundary conditions imposed on the model(top) and the 3D model as used in our finite element analysis(bottom).

**Table 3:** Materials parameters of the viscoelastic-viscoplastic damage model

|  | Parameter | Value | Equation | References |
|---|---|---|---|---|
| Equilibrium shear modulus | $\mu_{eq}^0$(MPa) | 760 | 15 |  |
| Non-equilibrium shear modulus | $\mu_{neq}^0$(MPa) | 790 | 16 |  |
| Volumetric bulk modulus | $\kappa_v$(MPa) | 1154 | 10 |  |
| Viscoelastic dashpot | $\dot{\varepsilon}_0$ (s$^{-1}$) | 1.0447 x 10$^{12}$ | 21 | [42] |
|  | $\Delta H$(J) | 1.977 x 10$^{-19}$ | 21 | [42] |
|  | $m$ | 0.657 | 21 |  |
|  | y$_0$ | 75 | 22 |  |
|  | x$_0$ | 0.2369 | 22 |  |
|  | b$_s$ | 0.06786 | 22 |  |
|  | a$_s$ | -48.23 | 22 |  |
| Viscoplastic dashpot | a | 0.179 | 27 |  |
|  | b | 0.910 | 27 |  |
|  | $\sigma_0$(MPa) | 5.5 | 27 |  |
| Damage | A | 320 | 13 |  |
| Moisture swelling coefficient | $a_w$ | 0.039 | 3 | [45] |

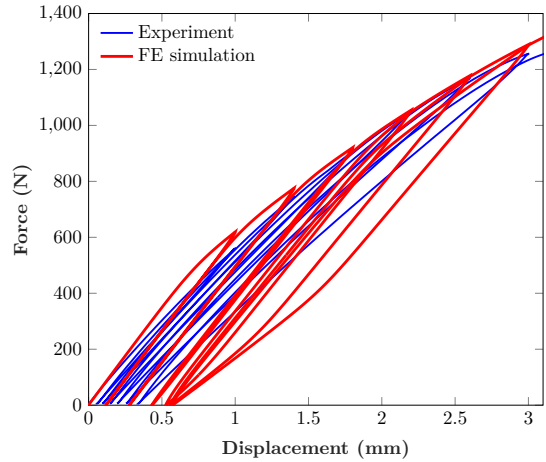

**Figure 12:** Force-displacement curve of dry epoxy system without BNPs at a load rate of 1 mm/min and room temperature under cyclic loading-unloading conditions.

*5.3. Training and validation of the deep learning model*

The trainable parameters $\omega$ for the LSTM units and the dense forward layer are trained using the mean average error (MAE) as the loss function for each batch size M and the number of features. The MAE is proven to be not as sensitive as the mean squared error regarding

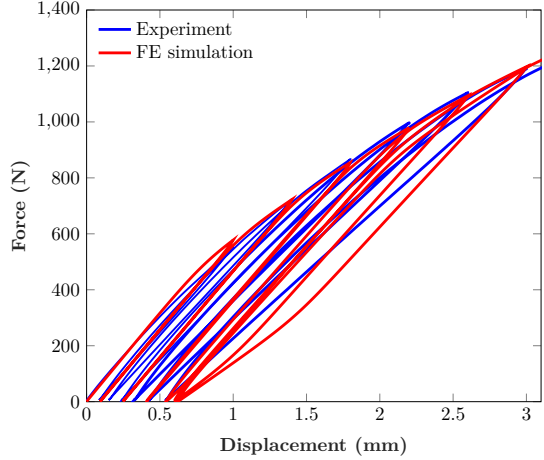

**Figure 13:** Experimental force-displacement response of the epoxy system without BNPs at saturated condition and finite element response obtained by the calibrated constitutive model at a load-rate of 1 mm/min.

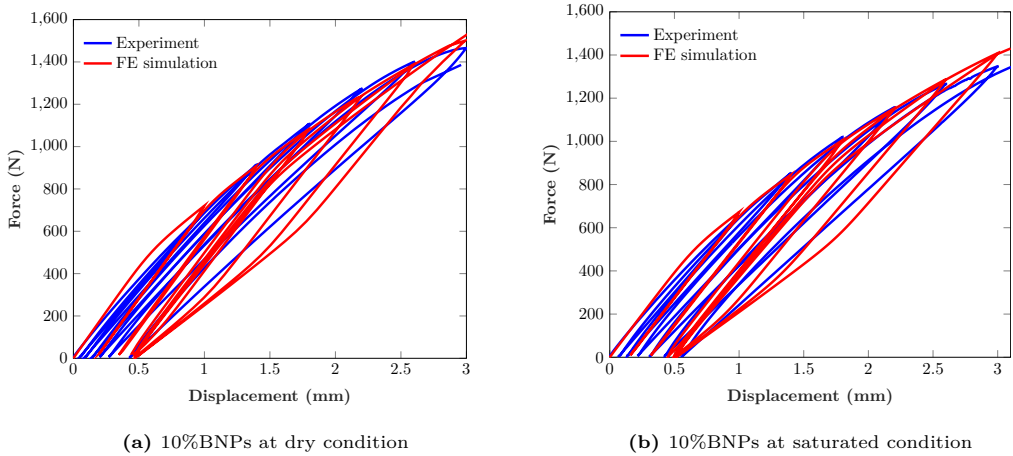

**(a)** 10%BNPs at dry condition

**(b)** 10%BNPs at saturated condition

**Figure 14:** Comparison of experimental and numerical force-displacement response of the epoxy/BNPs system for loading-unloading tests with increasing amplitude at dry/saturated conditions and a load-rate of 1 mm/min.

outliers [12]. The error $\epsilon$ is computed using the loss function, and the contribution of each trainable parameter is backpropagated to calculate the gradient and train the deep network. The optimization algorithm implemented is the Adam optimizer [66] with a learning rate of $\eta = 0.001$ which is dropped to $\eta = 0.0001$ after 200 epochs. The loss function adopts the L1 loss, yielding smooth results according to cross-validation.

Firstly, we tune the number of hidden units for both layers at a fixed batch size of M = 64.

The presented results of the training are shown in Fig. 15, and the mean average error is calculated as follows:

$$\text{MAE} = \frac{1}{N} \sum_{i=1}^{N} \left| \boldsymbol{\sigma}_{tot}^{pred} - \boldsymbol{\sigma}_{tot}^{targ} \right|, \tag{60}$$

where $\boldsymbol{\sigma}_{tot}^{pred}$ is the predicted value by the DL model at the i$th$ timestep of a sequence in the training data, $\boldsymbol{\sigma}_{tot}^{targ}$ is the correct corresponding value and N is the number of data. The mean(MAE) is calculated by taking the mean value of the last 50 epochs. It can be observed that the most accurate results are obtained by using 150 LSTM units. Consequently, we continue the training using 150 units for each layer, which is also preferable due to less trainable parameters and leads assumingly to a better performance regarding computational efficiency compared with a larger number of units.

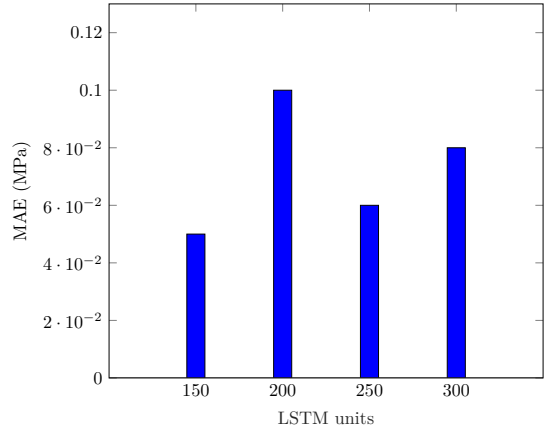

**Figure 15:** Performance of the two-layer LSTM architecture as a function of the number of LSTM units per layer.

Considering the choice of the batch size, representing the number of training examples in one backward pass, the DL model is trained on varying batch sizes of $[32, 64, 128]$. The results are presented in Table 4 and show that using a batch size of 64 leads to the best results in the training data. The final results using 150 LSTM units and a batch size of 64 are presented in Fig. 16. As can be seen, lowering the learning rate after 200 epochs helped the DL model to achieve an accurate result leading to a low plateau at a loss of $5 \cdot 10^{-2}$.

As mentioned above, the validation of the model is done using 10% of the generated data leading to a loss of $8 \cdot 10^{-2}$ MPa. Hence, the following results are obtained using validation data.

**Table 4:** Summary of the training data results for three different batch sizes

|  | Batch size | | |
|---|---|---|---|
|  | 32 | 64 | 128 |
| mean(MAE) | $8 \cdot 10^{-2}$ | $5 \cdot 10^{-2}$ | $7 \cdot 10^{-2}$ |

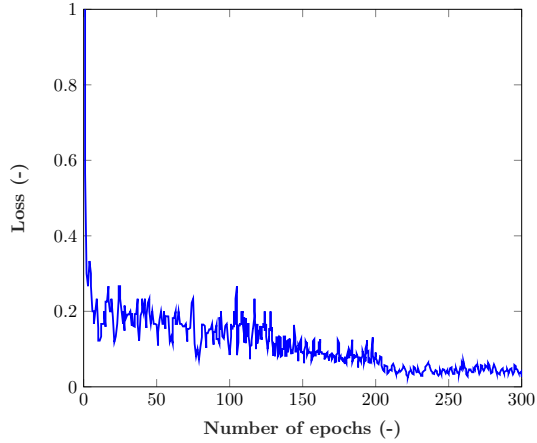

**Figure 16:** Training loss during the training of the DL model. The loss converges towards a plateau after 200 epochs.

Fig. 17(a) and Fig. 17(b) show the stress-strain behavior of BNP/epoxy nanocomposites under uniaxial cyclic loading predicted by the constitutive and DL model, respectively. A comparison between the simulation results reveals that the DL model is able to learn the nonlinear rate-dependent material behavior at both dry and saturated conditions. Fig. 18 also presents the behavior of the epoxy under a complex cyclic path of uniaxial and shear deformations. In this case, we visit a total of P = 5 points for each deformation gradient component in the 9-dimensional spatial space as presented in Table 1, resulting to an unique loading path for each direction and thus making the deformation scenario as complex as possible. As can be seen, the DL model can predict the stress tensor accurately under complex loading scenarios and can, therefore, replace the constitutive model.

Next, a set of uniaxial cyclic loading-unloading results is presented in Fig. 19 to evaluate the DL model's capability to predict the epoxy's rate-dependent behavior. As can be seen, the DL model is able to accurately predict the rate-dependent behavior in excellent agreement with the constitutive model for a wide range of strain rates. Furthermore, the imposed deformation

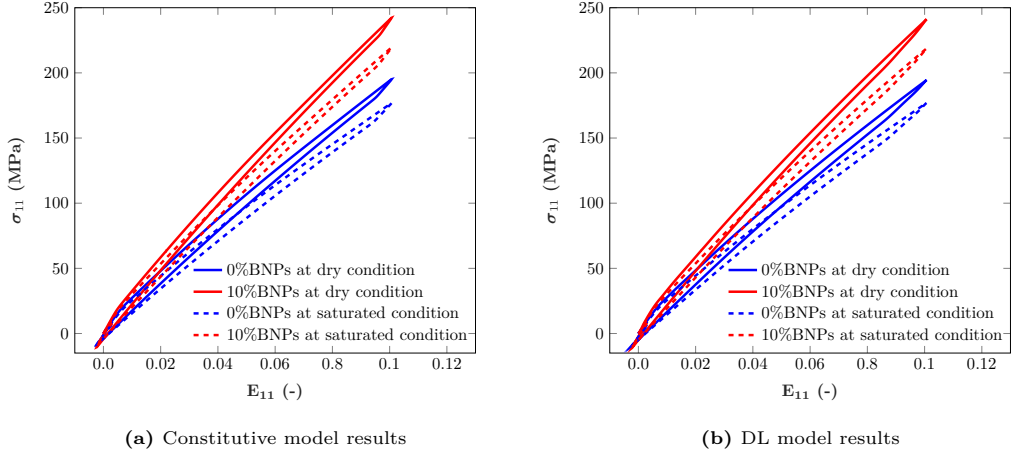

**(a)** Constitutive model results

**(b)** DL model results

**Figure 17:** Uniaxial loading results of the constitutive model and DL model for different amount of BNPs volume fraction at dry and moisture saturated state. $E_{11}$ represents the Green strain element and is obtained using Eq. (28). The strain rate is $\dot{\varepsilon} = 5 \times 10^{-4}\ s^{-1}$.

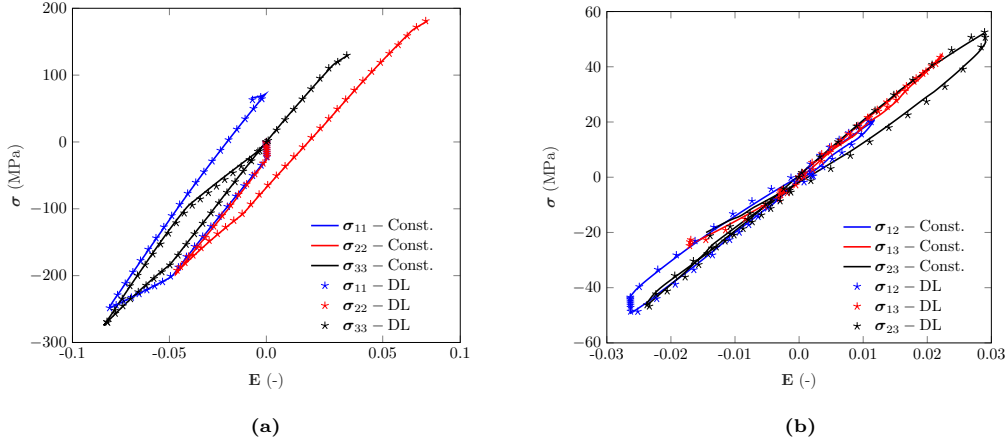

**(a)**

**(b)**

**Figure 18:** The neat and dry epoxy under a complex three-dimensional combination of shear and axial cyclic loading. The diagonal terms of the stress tensor are presented in (a) and the shear terms are shown in (b). The effective strain rate derived from the frobenius norm of the Green strain tensor is $\dot{\varepsilon}_F = 3 \times 10^{-4}\ s^{-1}$.

path includes compression, confirming the predictive ability regarding the transition of the stress state from tensile to compression.

### 5.4. Computational efficiency

In the following subsection, the computational efficiency of the DL model is compared with the constitutive model at the rheological level. As stated earlier, the Euler backward time integration scheme is used to formulate the constitutive model. Fig. 20 presents a deformation path with a deformation step of $\Delta F = 8.7 \cdot 10^{-5}$ and a timestep of $\Delta t = 1.28s$. It should be noted that the chosen timestep is not applicable for an explicit time integration since no convergence within the specified accuracy can be reached. The Euler backward integration scheme and the forward propagation of the DL model are implemented in an in-house MATLAB code. The simulation results show that the computational efficiency strongly depends on the deformation path's complexity. While simple deformation paths, such as uniaxial loading, lead to no or low acceleration of the model, a reduction of the computation time of the model with complex loading paths by a factor of 3.5 is detected. The path includes a combination of all deformation gradient components. It can be regarded as one of the most complex paths since it contains all deformation directions in the 3D stress-strain space. Table 5 shows the CPU-time needed for two combinations of BNPs volume fraction at dry and saturated state. We note a reduction of the computation time by a factor of 3.5 for 10% BNPs volume fraction at dry state, while other combinations indicate no or low accelerations. Another noteworthy observation is the constant CPU time of the DL model, which is expected as the forward propagation includes no iterative scheme, and only matrix multiplications are repeated in each loading step.

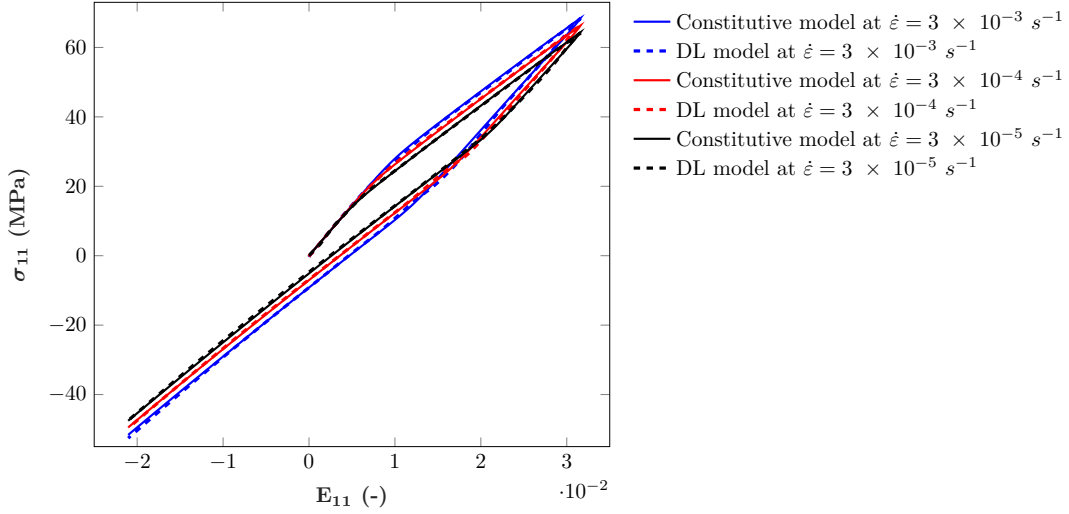

**Figure 19:** Comparison of the constitutive model against the DL model for the uniaxial case of the neat and saturated epoxy system under cyclic loading-unloading conditions at three different strain-rates.

This behavior is elucidated by observing the numerical time passed at each timestep within the Euler backward algorithm compared with the DL model. The results are presented in

**Table 5:** Comparison of the CPU time for the DL and constitutive model in seconds.

|  |  | BNPs volume fraction | |
|---|---|---|---|
|  |  | 0% | 10% |
| Dry | Constitutive model | 0.57 | 1.24 |
|  | DL model | 0.35 | 0.35 |
| Saturated | Constitutive model | 0.31 | 0.44 |
|  | DL model | 0.35 | 0.35 |

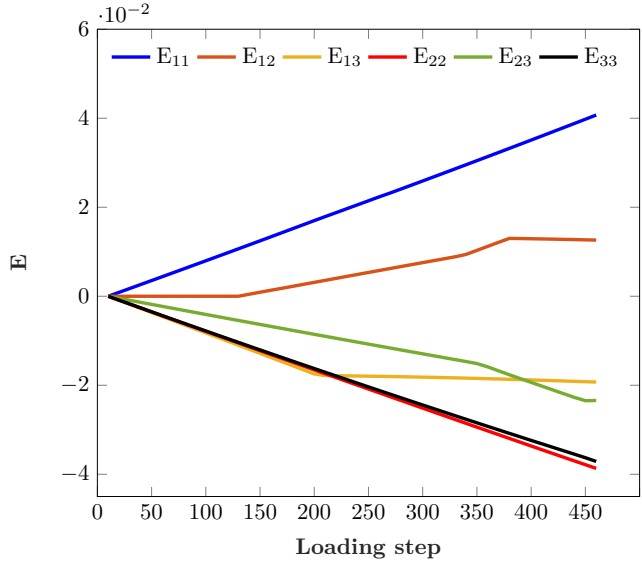

**Figure 20:** Exemplery loading path including all nine deformation gradient components presented using the Eq. (28) as Green strain components.

Fig. 21. In each combination, the Euler backward algorithm needs a low number of iterations at the beginning of each run, possibly due to the inactivity of the viscous or viscoplastic dashpot presented in Fig. 1. After approximately 300 steps, the integration time per step increases up to a peak of 0.28 s, which is an indication that the viscous and viscoplastic dashpots are activated, and the Euler backward algorithm requires multiple iterations to fully integrate the model within a tolerance of $1 \cdot 10^{-5}$. It also implies that an explicit time integration is not a reasonable choice for the chosen timestep. Also, a proportional increase in integration time per step is observed by increasing the model's stiffness. Accordingly, the simulation time per timestep for the DL model is constant at $7.7 \cdot 10^{-4}$s.

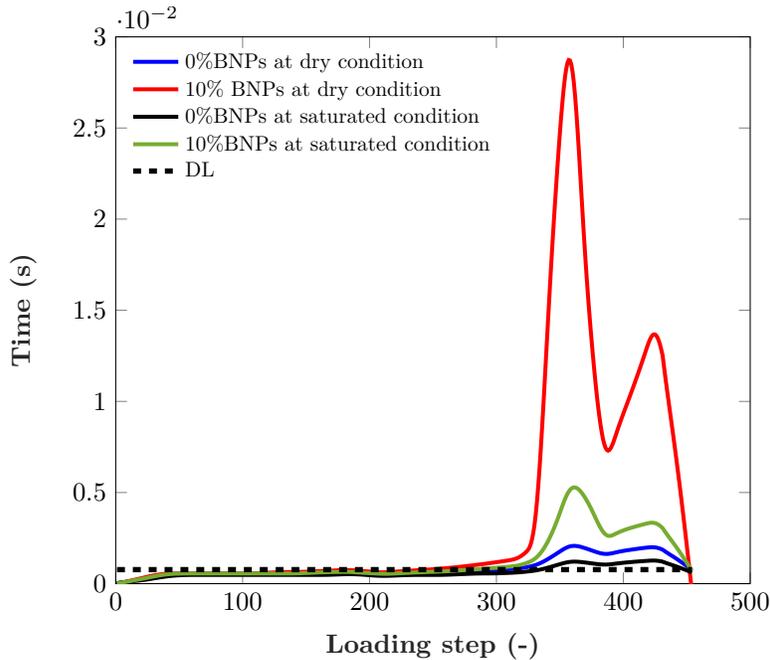

**Figure 21:** Time needed for each loading step within the Euler backward algorithm to integrate the constitutive model compared with the DL model.

*5.5. Accuracy of the proposed DL model*

The accuracy of the proposed DL model has already been verified in Section 5.3. Here, the accuracy of the trained and validated DL model within the FE framework is evaluated by comparing the results of the force-displacement curves for the 3D model presented in Fig. 11. FE simulations are performed by implementing the trained deep network within a FE code in C++ and utilizing DealII libraries [67] combined with the Eigen library for the matrix multiplication [68].

**Table 6:** MAE of the DL model compared to the constitutive model

|  | BNPs volume fraction | |
| --- | --- | --- |
|  | 0% | 10% |
| Dry | 20.9 | 5.9 |
| Saturated | 17.8 | 14.6 |

The obtained force-displacement curves based on the DL and constitutive models are presented in Fig. 22. For a better comparison, the MAE of the force-displacement curves is detailed in Table 6, suggesting a good agreement between simulation results obtained by the two material models. The FE simulation results confirm the predictive capability of the DL model in capturing the nonlinear behavior of BNP/epoxy nanocomposites. It should be mentioned that a converged solution within the Newton-Raphson method can only be reached with the proposed perturbation algorithm within the data generation described in algorithm 1. It implies that the stress-strain behavior can be characterized by generating data without the algorithm 1. Nevertheless, an approximation of the tangent modulus $\hat{\mathbb{C}}$ without the algorithm 1 is solely possible for single-element case studies and failed for the specific three-dimensional model presented above.

Conclusively, the agreement between the experimental data and FE facilitated by the DL model assures a significant increase in computational efficiency while only a marginal decrease in accuracy is observed.

Also, it is observed that the CPU time of FE simulations using the DL model is decreased by a factor of 1.5 for the model with 10% BNP volume fraction at the dry condition. Also, a reduction by a factor of 1.3 is noticed for the model without nanoparticles at the saturated condition. The simulations are run on 30 CPUs of Intel Cascade Lake Xeon Gold 6230N (2.3GHz, 30MB Cache, 125W) using multithreading. Although the simulation involves only uniaxial loading, the stress concentration at the middle of the specimen increases the computational costs for the constitutive model. Accordingly, the DL model runs at constant CPU time, reducing the wall time by a factor of 1.6 and 1.4, respectively.

## 6. Summary and conclusions

A nonlinear viscoelastic-viscoplastic damage model has been proposed to investigate the mechanical behavior of BNP/epoxy nanocomposites with moisture content at finite deformation. We implemented the Guth-Gold model to predict the impact of nanoparticles and moisture content on the material behavior. Also, the athermal yield stress related to the viscoelastic Argon model was modified by proposing a nonlinear sigmoid function and the chain stretch as the driving force. The results show that the proposed constitutive model is able to accurately predict the nonlinear cyclic behavior of the nanocomposite. To accelerate finite element simulations, a LSTM enhanced network was trained to predict the highly nonlinear material behavior of the nanocomposites. The traditional viscoelastic-viscoplastic model uti-lizing the Euler backward algorithm for integration of the constitutive equations was replaced with the trained DL model enhanced by two layers of LSTM units. We proposed a data generation framework using a space-filling approach and perturbing loading paths to obtain a reasonable set of data for the supervised learning of the DL model to predict the stress-strain behavior and to compute the consistent tangent modulus. Benchmarks examples show that the elimination of the iterative integration algorithm with the trained DL model leads to a significant increase of the computational efficiency and the DL model is capable to predict complex cyclic loading paths for a wide range of strain-rates. Also, the implemented approach to perturb the loading path within the data generation algorithm leads to an accurate computation of the consistent tangent modulus as needed in the finite element simulations. While the proposed data generation approach can be used for most of the constitutive models to fully capture the three-dimensional stress-strain behavior, a reasonable question is whether the generation of data and the training is worth the time. It should be noted here, that the speed up is especially useful if we are interested in complex constitutive models as presented here, where the numerical integration is a significant part of the computational cost. Thus, employing a DL model for simple constitutive models, e.g. linear elasticity, can lead to an increase of the computational cost. Also, the proposed DL model can be further extended to adapt to different materials by using additional inputs e.g. material parameters. This can be easily achieved, since the generated deformation paths are saved and we can rerun the algorithm in Table 1 starting from STEP 6 to generate the output for a different material. Another feature is that the model can be also extended to adapt to different ambient conditions like temperature dependency, making the model as universal as possible. Since the constitutive model does already include the temperature dependency within the viscoelastic Argon model as presented in Eq. (21), we can generate additional training data to capture this dependency as well without the need to change the constitutive model at hand.

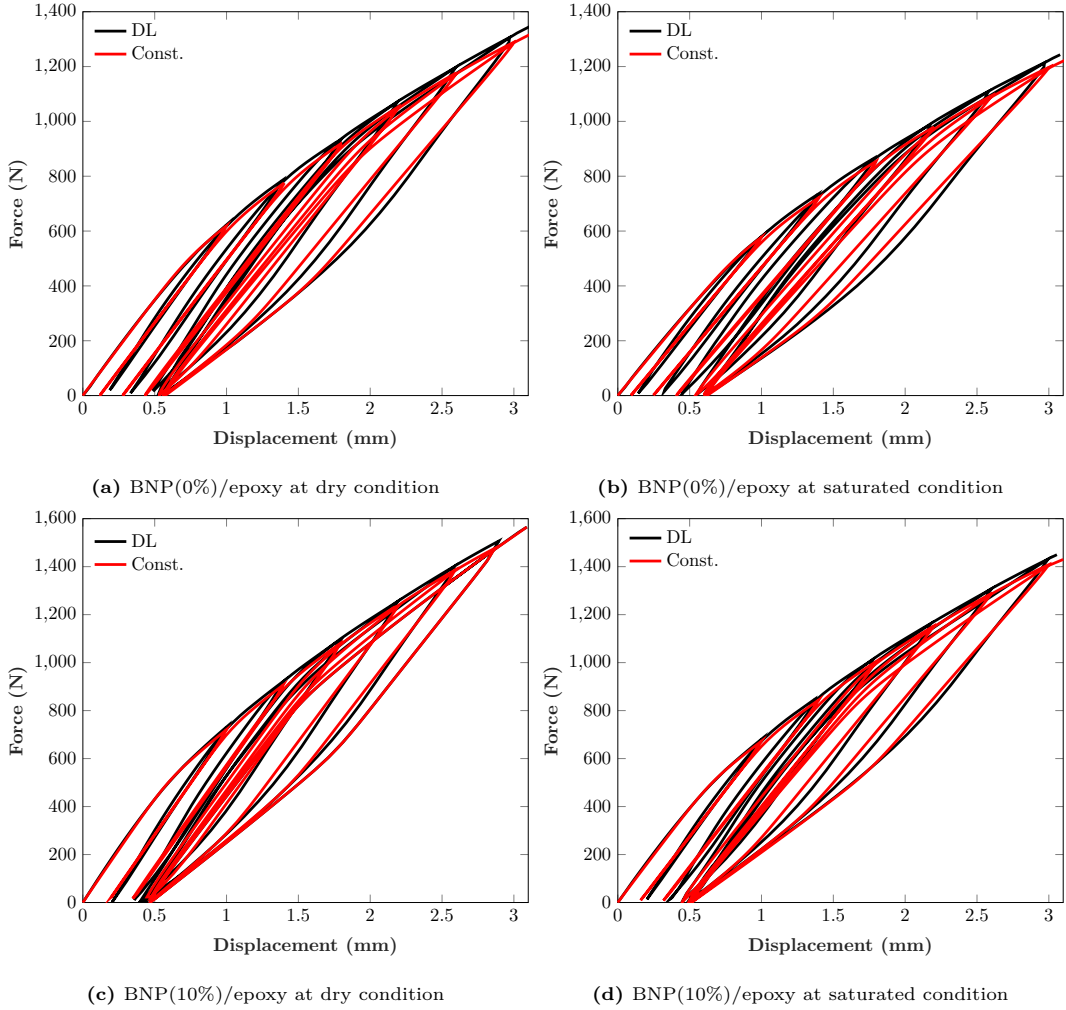

**Figure 22:** Force-displacement response for the DL- and constitutive model associated with specimens made of BNP/epoxy at dry and saturated conditions.

Since the constitutive model is implemented in FE analysis in a modular fashion, integrating the DL model was straightforward. It led to a remarkable increase in the computational efficiency of FE analysis, while only a marginal decrease in accuracy is observed.

In summary, the proposed DL model can be further developed to integrate other ambient conditions, such as temperature dependency, and should be compared with experimental data. Furthermore, in future studies, the effect of non-uniform dispersion of nanoparticles and moisture content at different temperatures can also be introduced within the developed model to increase its flexibility.

*Data availability*

The source codes of the finite element analysis in this work are available at `https://github.com/BBahtiri/viscoelastic_viscoplastic_model`.

*Acknowledgements*

This work originates from the following research project: "Challenges of industrial application of nanomodified and hybrid material systems in lightweight rotor blade construction" ("HANNAH - Herausforderungen der industriellen Anwendung von nanomodifizierten und hybriden Werkstoffsystemen im Rotorblattleichtbau"), funded by the Federal Ministry for Economic Affairs and Energy, Germany. The authors wish to express their gratitude for the financial support. The authors acknowledge the support of the LUIS scientific computing cluster, Germany, which is funded by Leibniz Universitat Hannover, Germany, the Lower Saxony Ministry of Science and Culture (MWK), Germany and the German Research Council (DFG).

**References**

[1] Y. Li, S. Wang, B. Arash, Q. Wang, A study on tribology of nitrile-butadiene rubber composites by incorporation of carbon nanotubes: Molecular dynamics simulations, Carbon 100 (2016) 145–150.

[2] B. Arash, W. Exner, R. Rolfes, A viscoelastic damage model for nanoparticle/epoxy nanocomposites at finite strain: A multiscale approach, Journal of the Mechanics and Physics of Solids 128 (2019) 162–180.

[3] B. Arash, W. Exner, R. Rolfes, Effect of moisture on the nonlinear viscoelastic fracture behavior of polymer nanocompsites: a finite deformation phase-field model, Engineering with Computers (2022) 1–18.

[4] M. Stoffel, R. Gulakala, F. Bamer, B. Markert, Artificial neural networks in structural dynamics: A new modular radial basis function approach vs. convolutional and feedforward topologies, Computer Methods in Applied Mechanics and Engineering 364 (2020) 112989.

[5] X. Liu, S. Tian, F. Tao, W. Yu, A review of artificial neural networks in the constitutive modeling of composite materials, Composites Part B: Engineering 224 (2021) 109152.

[6] F. Pled, C. Desceliers, T. Zhang, A robust solution of a statistical inverse problem in multiscale computational mechanics using an artificial neural network, Computer Methods in Applied Mechanics and Engineering 373 (2021) 113540.

[7] P. Pantidis, M. E. Mobasher, Integrated finite element neural network (i-fenn) for nonlocal continuum damage mechanics, Computer Methods in Applied Mechanics and Engineering 404 (2023) 115766.

[8] F. Tao, X. Liu, H. Du, W. Yu, Finite element coupled positive definite deep neural networks mechanics system for constitutive modeling of composites, Computer Methods in Applied Mechanics and Engineering 391 (2022) 114548.

[9] J. Ghaboussi, J. Garrett Jr, X. Wu, Knowledge-based modeling of material behavior with neural networks, Journal of engineering mechanics 117 (1) (1991) 132–153.

[10] J. Ghaboussi, D. A. Pecknold, M. Zhang, R. M. Haj-Ali, Autoprogressive training of neural network constitutive models, International Journal for Numerical Methods in Engineering 42 (1) (1998) 105–126.

[11] M. Stoffel, F. Bamer, B. Markert, Neural network based constitutive modeling of nonlinear viscoplastic structural response, Mechanics Research Communications 95 (2019) 85–88.

[12] C. Wang, L.-y. Xu, J.-s. Fan, A general deep learning framework for history-dependent response prediction based on ua-seq2seq model, Computer Methods in Applied Mechanics and Engineering 372 (2020) 113357.

[13] F. Feyel, A multilevel finite element method (fe2) to describe the response of highly nonlinear structures using generalized continua, Computer Methods in applied Mechanics and engineering 192 (28-30) (2003) 3233–3244.

[14] E. M. Haghighi, S. Na, A single long short-term memory network for enhancing the prediction of path-dependent plasticity with material heterogeneity and anisotropy, arXiv preprint arXiv:2204.01466 (2022).

[15] F. Ghavamian, A. Simone, Accelerating multiscale finite element simulations of history-dependent materials using a recurrent neural network, Computer Methods in Applied Mechanics and Engineering 357 (2019) 112594.

[16] D. E. Rumelhart, G. E. Hinton, R. J. Williams, Learning representations by back-propagating errors, nature 323 (6088) (1986) 533–536.

[17] J. Chung, C. Gulcehre, K. Cho, Y. Bengio, Empirical evaluation of gated recurrent neural networks on sequence modeling, arXiv preprint arXiv:1412.3555 (2014).

[18] M. Mundt, S. David, A. Koeppe, F. Bamer, B. Markert, W. Potthast, Intelligent prediction of kinetic parameters during cutting manoeuvres, Medical & Biological Engineering & Computing 57 (2019) 1833–1841.

[19] Learning to forget: Continual prediction with lstm.

[20] S. Hochreiter, J. Schmidhuber, Long short-term memory, Neural computation 9 (8) (1997) 1735–1780.

[21] I. Goodfellow, Y. Bengio, A. Courville, Deep learning, MIT press, 2016.

[22] A. Koeppe, F. Bamer, M. Selzer, B. Nestler, B. Markert, Explainable artificial intelligence for mechanics: physics-explaining neural networks for constitutive models, Frontiers in Materials 8 (2022) 636.

[23] L. Wu, N. G. Kilingar, L. Noels, et al., A recurrent neural network-accelerated multi-scale model for elasto-plastic heterogeneous materials subjected to random cyclic and non-proportional loading paths, Computer Methods in Applied Mechanics and Engineering 369 (2020) 113234.

[24] L. Benabou, Development of lstm networks for predicting viscoplasticity with effects of deformation, strain rate, and temperature history, Journal of Applied Mechanics 88 (7) (2021).

[25] M. H. Sadeghi, S. Lotfan, Identification of non-linear parameter of a cantilever beam model with boundary condition non-linearity in the presence of noise: an nsi-and ann-based approach, Acta Mechanica 228 (2017) 4451–4469.

[26] A. Koeppe, F. Bamer, B. Markert, An intelligent nonlinear meta element for elasto-plastic continua: deep learning using a new time-distributed residual u-net architecture, Computer Methods in Applied Mechanics and Engineering 366 (2020) 113088.

[27] S. B. Tandale, F. Bamer, B. Markert, M. Stoffel, Physics-based self-learning recurrent neural network enhanced time integration scheme for computing viscoplastic structural finite element response, Computer Methods in Applied Mechanics and Engineering 401 (2022) 115668.

[28] R. Arora, P. Kakkar, B. Dey, A. Chakraborty, Physics-informed neural networks for modeling rate-and temperature-dependent plasticity, arXiv preprint arXiv:2201.08363 (2022).

[29] R. Tipireddy, P. Perdikaris, P. Stinis, A. Tartakovsky, A comparative study of physics-informed neural network models for learning unknown dynamics and constitutive relations, arXiv preprint arXiv:1904.04058 (2019).

[30] E. Zhang, M. Yin, G. E. Karniadakis, Physics-informed neural networks for nonhomogeneous material identification in elasticity imaging, arXiv preprint arXiv:2009.04525 (2020).

[31] R. Arora, Machine learning-accelerated computational solid mechanics: Application to linear elasticity, arXiv preprint arXiv:2112.08676 (2021).

[32] M. Jux, J. Fankhanel, B. Daum, T. Mahrholz, M. Sinapius, R. Rolfes, Mechanical properties of epoxy/boehmite nanocomposites in dependency of mass fraction and surface modification-an experimental and numerical approach, Polymer 141 (2018) 34–45.

[33] V.-D. Nguyen, F. Lani, T. Pardoen, X. Morelle, L. Noels, A large strain hyperelastic viscoelastic-viscoplastic-damage constitutive model based on a multi-mechanism non-local damage continuum for amorphous glassy polymers, International Journal of Solids and Structures 96 (2016) 192–216.

[34] E. Kontou, Viscoplastic deformation of an epoxy resin at elevated temperatures, Journal of applied polymer science 101 (3) (2006) 2027–2033.

[35] P. Yu, X. Yao, Q. Han, S. Zang, Y. Gu, A visco-elastoplastic constitutive model for large deformation response of polycarbonate over a wide range of strain rates and temperatures, Polymer 55 (25) (2014) 6577–6593.

[36] M. N. Silberstein, M. C. Boyce, Constitutive modeling of the rate, temperature, and hydration dependent deformation response of nafion to monotonic and cyclic loading, Journal of Power Sources 195 (17) (2010) 5692–5706.

[37] H. J. Qi, M. C. Boyce, Stress–strain behavior of thermoplastic polyurethanes, Mechanics of materials 37 (8) (2005) 817–839.

[38] A. Melro, P. Camanho, F. A. Pires, S. Pinho, Micromechanical analysis of polymer composites reinforced by unidirectional fibres: Part i–constitutive modelling, International Journal of Solids and Structures 50 (11-12) (2013) 1897–1905.

[39] X. Poulain, A. Benzerga, R. Goldberg, Finite-strain elasto-viscoplastic behavior of an epoxy resin: Experiments and modeling in the glassy regime, International Journal of Plasticity 62 (2014) 138–161.

[40] J. Sweeney, I. M. Ward, Mechanical properties of solid polymers, John Wiley & Sons, 2012.

[41] B. Bahtiri, B. Arash, R. Rolfes, Elucidating atomistic mechanisms underlying water diffusion in amorphous polymers: An autonomous basin climbing-based simulation method, Computational Materials Science 212 (2022) 111565.

[42] R. Unger, W. Exner, B. Arash, R. Rolfes, Non-linear viscoelasticity of epoxy resins: Molecular simulation-based prediction and experimental validation, Polymer 180 (2019) 121722.

[43] I. Rocha, F. van der Meer, S. Raijmaekers, F. Lahuerta, R. Nijssen, L. J. Sluys, Numerical/experimental study of the monotonic and cyclic viscoelastic/viscoplastic/fracture behavior of an epoxy resin, International Journal of Solids and Structures 168 (2019) 153–165.

[44] J. Bergstrom, L. Hilbert Jr, A constitutive model for predicting the large deformation thermomechanical behavior of fluoropolymers, Mechanics of Materials 37 (8) (2005) 899–913.

[45] T. Cui, P. Verberne, S. Meguid, Characterization and atomistic modeling of the effect of water absorption on the mechanical properties of thermoset polymers, Acta Mechanica 229 (2) (2018) 745–761.

[46] L. Mullins, N. Tobin, Stress softening in rubber vulcanizates. part i. use of a strain amplification factor to describe the elastic behavior of filler-reinforced vulcanized rubber, Journal of Applied Polymer Science 9 (9) (1965) 2993–3009.

[47] L. Mullins, Softening of rubber by deformation, Rubber chemistry and technology 42 (1) (1969) 339–362.

[48] S. Govindjee, S. Reese, A presentation and comparison of two large deformation viscoelasticity models (1997).

[49] E. Guth, Theory of filler reinforcement, Rubber Chemistry and Technology 18 (3) (1945) 596–604.

[50] K. Ho, E. Krempl, Extension of the viscoplasticity theory based on overstress (vbo) to capture non-standard rate dependence in solids, International Journal of Plasticity 18 (7) (2002) 851–872.

[51] Z. Xia, Y. Hu, F. Ellyin, Deformation behavior of an epoxy resin subject to multiaxial loadings. part ii: Constitutive modeling and predictions, Polymer Engineering & Science 43 (3) (2003) 734–748.

[52] J. Bergstrom, S. Kurtz, C. Rimnac, A. Edidin, Constitutive modeling of ultra-high molecular weight polyethylene under large-deformation and cyclic loading conditions, Biomaterials 23 (11) (2002) 2329–2343.

[53] J. N. Fuhg, N. Bouklas, On physics-informed data-driven isotropic and anisotropic constitutive models through probabilistic machine learning and space-filling sampling, Computer Methods in Applied Mechanics and Engineering 394 (2022) 114915.

[54] D. Penumadu, R. Zhao, Triaxial compression behavior of sand and gravel using artificial neural networks (ann), Computers and Geotechnics 24 (3) (1999) 207–230.

[55] V. Papadopoulos, G. Soimiris, D. Giovanis, M. Papadrakakis, A neural network-based surrogate model for carbon nanotubes with geometric nonlinearities, Computer Methods in Applied Mechanics and Engineering 328 (2018) 411–430.

[56] M. Lefik, D. Boso, B. Schrefler, Artificial neural networks in numerical modelling of composites, Computer Methods in Applied Mechanics and Engineering 198 (21-26) (2009) 1785–1804.

[57] A. Dekhovich, O. T. Turan, J. Yi, M. A. Bessa, Cooperative data-driven modeling, arXiv preprint arXiv:2211.12971 (2022).

[58] H. Kim, I. Jeong, H. Cho, M. Cho, Surrogate model based on data-driven model reduction for inelastic behavior of composite microstructure, International Journal of Aeronautical and Space Sciences (2022) 1–21.

[59] Y. Heider, K. Wang, W. Sun, So (3)-invariance of informed-graph-based deep neural network for anisotropic elastoplastic materials, Computer Methods in Applied Mechanics and Engineering 363 (2020) 112875.

[60] L. Wu, K. Zulueta, Z. Major, A. Arriaga, L. Noels, Bayesian inference of non-linear multiscale model parameters accelerated by a deep neural network, Computer Methods in Applied Mechanics and Engineering 360 (2020) 112693.

[61] L. Kocis, W. J. Whiten, Computational investigations of low-discrepancy sequences, ACM Transactions on Mathematical Software (TOMS) 23 (2) (1997) 266–294.

[62] Y. Hashash, S. Jung, J. Ghaboussi, Numerical implementation of a neural network based material model in finite element analysis, International Journal for numerical methods in engineering 59 (7) (2004) 989–1005.

[63] K. A. Kalina, L. Linden, J. Brummund, P. Metsch, M. Kastner, Automated constitutive modeling of isotropic hyperelasticity based on artificial neural networks, Computational Mechanics (2022) 1–20.

[64] P. Wriggers, Nonlinear finite element methods, Springer Science & Business Media, 2008.

[65] W. Sun, E. L. Chaikof, M. E. Levenston, Numerical approximation of tangent moduli for finite element implementations of nonlinear hyperelastic material models, Journal of biomechanical engineering 130 (6) (2008).

[66] D. P. Kingma, J. Ba, Adam: A method for stochastic optimization, arXiv preprint arXiv:1412.6980 (2014).

[67] D. Arndt, W. Bangerth, M. Feder, M. Fehling, R. Gassmoller, T. Heister, L. Heltai, M. Kronbichler, M. Maier, P. Munch, J.-P. Pelteret, S. Sticko, B. Turcksin, D. Wells, The `deal.II` library, version 9.4, Journal of Numerical Mathematics 30 (3) (2022) 231–246. `doi:10.1515/jnma-2022-0054`.
URL `https://dealii.org/deal94-preprint.pdf`

[68] G. Guennebaud, B. Jacob, et al., Eigen v3, http://eigen.tuxfamily.org (2010).